\documentclass{isprs}
\usepackage{subfigure}
\usepackage{setspace}
\usepackage{geometry} 
\usepackage{enumerate}
\usepackage{natbib}
\usepackage{multirow}  
\usepackage{stfloats}
\usepackage{url}
\usepackage{amsmath}
\usepackage{amssymb}

\usepackage{float}
\usepackage{color}
\usepackage{soul}
\usepackage{algorithm}
\usepackage{diagbox}
\newcommand{\ourname}{AETree}

\begin{document}

\title{AutoEncoding Tree for City Generation and Applications}
\author{
Wenyu Han\textsuperscript{a}, Congcong Wen\textsuperscript{a}\thanks{Corresponding author. Email: wencc@nyu.edu. }, Lazarus Chok\textsuperscript{a}, Yan Liang Tan\textsuperscript{a}, Sheung Lung Chan\textsuperscript{a}, Hang Zhao\textsuperscript{b}, Chen Feng\textsuperscript{a}
}

\address
{
\textsuperscript{a }Tandon School of Engineering, New York University, New York, United States.\\
\textsuperscript{b }
Institute for Interdisciplinary Information Sciences, Tsinghua University, Beijing, China.\\
}

\abstract{
City modeling and generation have attracted an increased interest in various applications, including gaming, urban planning, and autonomous driving. Unlike previous works focused on the generation of single objects or indoor scenes, the huge volumes of spatial data in cities pose a challenge to the generative models. Furthermore, few publicly available 3D real-world city datasets also hinder the development of methods for city generation. In this paper, we first collect over 3,000,000 geo-referenced objects for the city of New York, Zurich, Tokyo, Berlin, Boston and several other large cities. Based on this dataset, we propose AETree, a tree-structured auto-encoder neural network, for city generation. Specifically, we first propose a novel Spatial-Geometric Distance (SGD) metric to measure the similarity between building layouts and then construct a binary tree  over the raw geometric data of building based on the SGD metric. Next, we present a tree-structured network whose encoder learns to extract and merge spatial information from bottom-up iteratively. The resulting global representation is reversely decoded for reconstruction or generation. To address the issue of long-dependency as the level of the tree increases, a Long Short-Term Memory (LSTM) Cell is employed as a basic network element of the proposed AETree. Moreover, we introduce a novel metric, Overlapping Area Ratio (OAR), to quantitatively evaluate the generation results. Experiments on the collected dataset demonstrate the effectiveness of the proposed model on 2D and 3D city generation. Furthermore, the latent features learned by AETree can serve downstream urban planning applications.}


\maketitle
\section{Introduction}\label{Introduction}
\begin{figure*}[t]
    \centering
\includegraphics[width=1\textwidth]{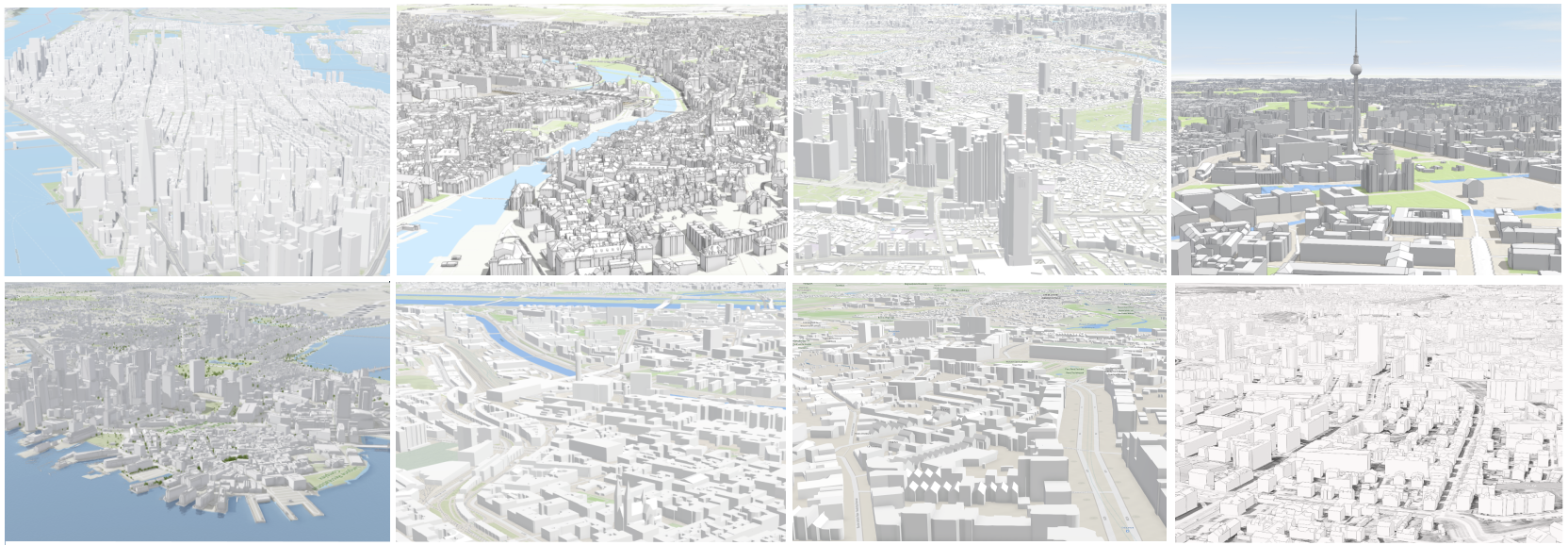}
    \caption{Overview of New York, Zurich, Tokyo, Berlin, Boston, Vienna, Brussels, and Estonia cities (order from top left to bottom right ) in our RealCity3D dataset.}
    \label{purity_example}
\end{figure*}



As cities represent a crucial domain for human activities, the have garnered significant attention in research. With the rapid advancements in image and video generation technologies, data-driven city generation has become increasingly attractive and feasible. This is primarily due to two key factors: the availability of abundant real-world data and the growing demand for such data in various applications. Notably, urban planners and architects heavily rely on city-level simulations to inform their designs. Game developers utilize city generation tools to autonomously generate virtual urban environments. In addition, the autonomous driving industry has seen a rising need for simulated environments, featuring novel maps and scenarios, to conduct road testing. These applications collectively drive the demand for realistic city generation techniques.


Despite the success of deep generative models in various data modalities, such as language, audio, images, videos, and even point clouds, several challenges hinder their application in city-level geometric generation. Firstly, cities consist of a complex array of geometrically parameterized objects with irregular layouts. Secondly, these objects typically exist within high-dimensional complex data manifolds. For instance, a building is represented as a collection of 3D polygons, each comprising a variable number of 3D vertices. Compared to LiDAR point clouds and remote sensing images, this data format is simpler and more lightweight, which can accelerate algorithm processing speed and reduce computational resources. Furthermore, there is a scarcity of publicly available 3D real-world city datasets that are well-suited for the development of data-driven deep generative models for cities.

Presently, existing public datasets for geometric data generation are categorized at different levels, including object-level~\citep{chang2015shapenet}, room-level~\citep{LIFULL,wu2019data}, and city-level~\citep{chu2019neural,spacenetdataset}. However, due to the limited scale of object-level and room-level datasets, training models on them to achieve city-level generation is challenging. Moreover, existing city-level datasets solely encompass 2D data, such as polylines and polygons, rendering it unrealistic to train models for 3D city generation using this 2D data. Consequently, some researchers~\citep{chang2021building} have created their synthetic 3D city-level datasets to facilitate building generation. Clearly, there is an urgent need for the proposal of a real-world 3D city-level dataset to enable more comprehensive research on deep generative models for city generation.

On the other hand, existing methods for urban generation can primarily be categorized into two main approaches: non-data-driven and data-driven methods. CityEngine, as a representative procedural modeling technique within the non-data-driven category, creates 3D city models based on manually defined grammars and rules. However, such methods require domain expertise and necessitate manual adjustments to the rules to accommodate new data. The other category, based on data-driven methods, typically employs deep learning for city generation. However, research in this field is relatively limited, with most deep learning-based generation work focusing on the object and indoor room levels, making it challenging to directly extend these approaches to large-scale city generation tasks. While some efforts have explored city-level generation, these works are primarily focus on generating 2D city layouts and often overlook the hierarchical spatial structures inherent in urban data.

In this paper, we introduce the RealCity3D dataset, a comprehensive 3D urban shape dataset encompassing cities including New York, Zurich, Tokyo, Berlin, Boston, Vienna, Brussels, and Estonia (Figure \ref{purity_example}). RealCity3D comprises an extensive collection of over 3,000,000 georeferenced objects, spanning more than 3,510 square kilometers of land area. These objects are conveniently available in four distinct formats: polygon mesh, triangle mesh, point cloud, and voxel grid. Crucially, semantic information pertaining to these objects is meticulously preserved within the polygon meshes. Leveraging the RealCity3D dataset, we propose AETree, a streamlined yet highly effective Auto-Encoder Tree neural network designed for city generation.  Specifically,  we first design a spatial-geometric distance metric and construct a binary tree from the original set of city buildings based on this metric. And then we present a tree-based encoder and decoder to hierarchically extract spatial information in a bottom-up manner and reconstruct the input data, respectively. To capture the long-dependencies among the different levels of the tree neural network, we incorporate the Long Short-Term Memory (LSTM) Cell as a basic network element in AETree. Furthermore, for a more comprehensive evaluation of city generation performance, we devise a novel metric, the Overlapping Area Ratio (OAR), designed to quantify the degree of overlap between generated buildings. Our main contributions can be summarized as follows:

\begin{itemize}

\item  We have open-sourced RealCity3D, a large-scale georeferenced 3D shape dataset, which includes over 3,000,000 georeferenced objects from 8 cities, covering an extensive land area of over 3,510 square kilometers. This dataset is available in various formats, including polygon meshes, triangular meshes, point clouds, and voxel grids. The dataset webpage is: \url{https://github.com/ai4ce/RealCity3D}.

\item  We introduce a novel distance metric called the Spatial Geometric Distance (SGD), which measures the similarity between physical objects in terms of both their spatial location and geometric properties.

\item We propose a simple yet effective tree auto-encoder neural network tailored for city generation. It employs the tree-based encoder and decoder to hierarchically extract spatial information, enhanced by LSTM Cells to capture long-range dependencies across different network levels.

\item We design a novel evaluation metric, the Overlapping Area Ratio (OAR), for city generation performance. This metric quantifies the degree of overlap between generated buildings, providing a more comprehensive assessment of the quality of city generation algorithms.



\end{itemize}

\section{Related work}

\begin{table*}[h]
\caption{Comparison with representative datasets}
	\label{table_compare_other_dataset}
	\resizebox{1\textwidth}{!}{%
\begin{tabular}{ccccccc}
\hline
 & Datasets & Year & Spatial extent & Objects & Format &  Generation Task \\ \hline
Object-level & ShapeNet~\citep{chang2015shapenet} & 2015 & - & 3,000,000+ &  Mesh & 3D Object Generation \\ \hline
\multirow{2}{*}{Room-level} & LIFULL HOME~\citep{LIFULL} & 2015 & - & 5,300,000+ &  Imagery & 2D Indoor Layout Generation \\
 & RPLAN~\citep{wu2019data} & 2019 & - & 80,000+ & Imagery & 2D Indoor Layout Generation \\ \hline
 
\multirow{6}{*}{City-level} 
& NYC3D~\citep{NYCdata_web} & 2014 & $3.02 \times 10^8 m^2$ & 850,000+ & CAD & 3D Object/City Generation\\
& RoadNet~\citep{chu2019neural} & 2019 & $1.7  \times 10^8 m^2 $ & - &  Polylines & 2D Road Network Generation \\
 & SpaceNet v2~\citep{vanetten2019spacenet} & 2018 & $30.11 \times 10^8 m^2$ & 685,000 & Imagery/Polygons & 2D City Layout Generation \\
 
& HoliCity~\citep{zhou2020holicity} & 2020 & $2 \times 10^7 m^2$ & - & CAD /Imagery & 3D Object Generation\\

& SUM~\citep{Gao_2021} & 2021 & $4 \times 10^6 m^2$ & - & Mesh & 3D Object Generation\\

& UrbanScene3D~\citep{lin2022capturing} & 2022 & $1.36 \times 10^8 m^2$ &  4,000,000+ & CAD/Mesh/Imagery & 3D Object/City Generation\\

 & RealCity3D & 2023 & $3.51 \times 10^9 m^2$  & 3,000,000+ &  Mesh/Point Cloud/Voxels & 3D Object and 2D/3D City Generation \\  
 \hline
\end{tabular}}
\end{table*}

\begin{figure*}[h] 
\centering
\includegraphics[width=0.7\textwidth]{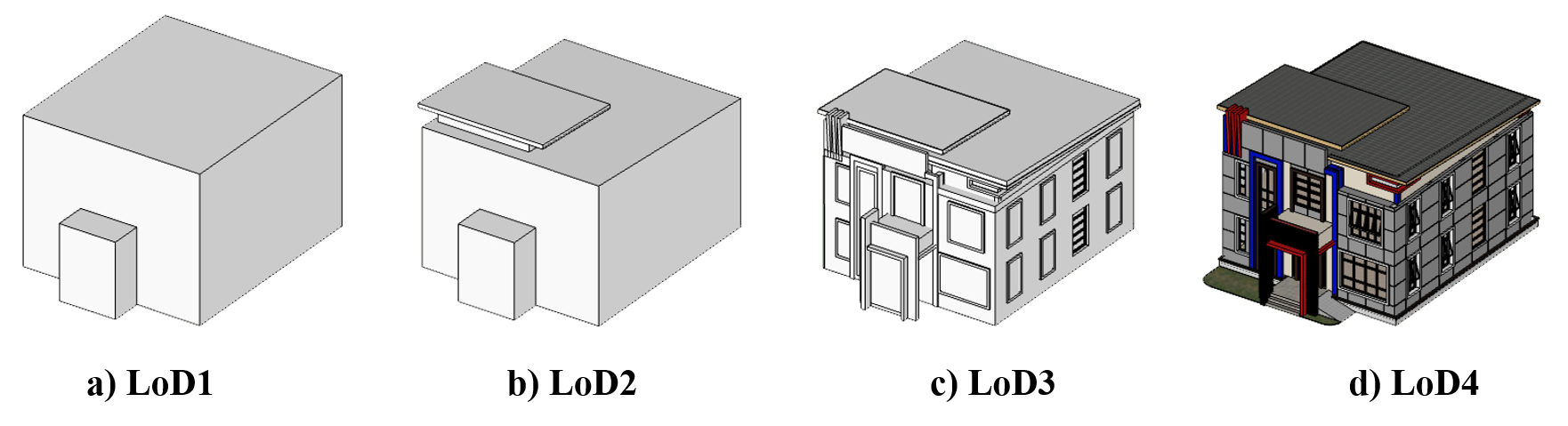}
\caption{
 Examples of the 3D building object ranging from LoD1 to LoD4. LoD4 sketch retrieved from SketchUp 3D Warehouse.
}
\label{fig2_citygml}
\end{figure*}

\begin{figure*}[hbp]
\begin{center}
\includegraphics[width=1\textwidth]{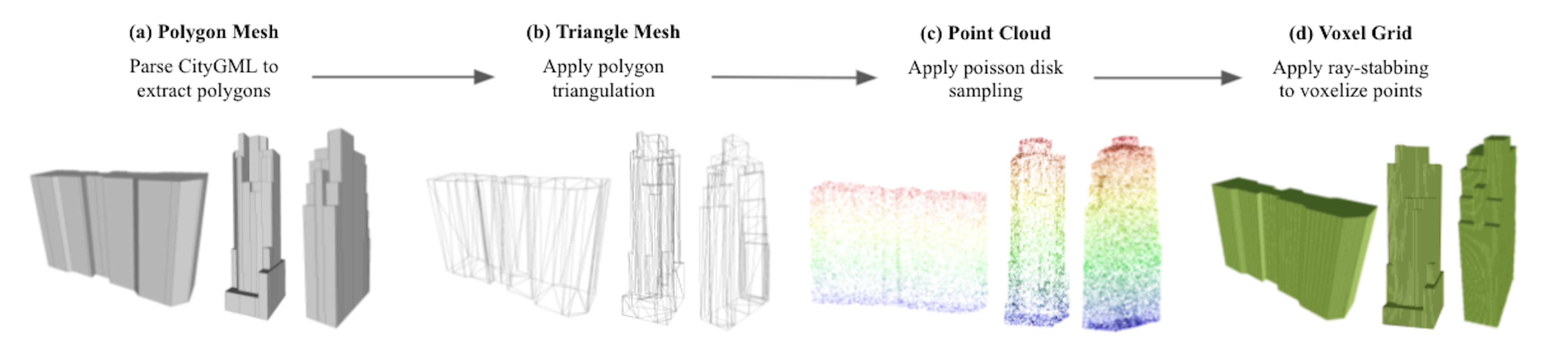}
\end{center}
\caption{Overview of RealCity3D dataset and transformation on a  LoD2 building shapes. (a) Polygon mesh and (b) triangle mesh  visualised in MeshLab, (c) point cloud in CloudCompare with colorized z-values, (d) voxel grid in viewvox.}
\label{fig3_flow_processing}
\end{figure*}

 \subsection{Datasets for Geometric Generation}
 
Geometric data generation methods have achieved great success in both research and real applications in vision, graphics, and robotic area.  Thank the recent public dataset (see Table~\ref{table_compare_other_dataset} for comparison) with three categories according to its scale:

\textbf{1) Object-level Datasets}: ShapeNet~\citep{chang2015shapenet} contains over three million 3D models with a core dataset of about 51,300 unique 3D models across 55 common object categories. Though some studies~\citep{groueix2018papier, achlioptas2018learning, nash2020polygen} achieve promising single object generation performance,  it will be difficult to extend trained models on this synthetic dataset to a large-scale and real-world dataset. On the contrary, RealCity3D deals exclusively with city-scale building objects georeferenced to the real world. 

\textbf{2) Room-level Datasets}: LIFULL HOME’s dataset~\citep{LIFULL} contains five million floor plans and RPLAN~\citep{wu2019data} consists of 80,000 floor plans from real-world residential buildings. Accordingly, Graph2Plan~\citep{hu2020graph2plan} and HouseGAN~\citep{nauata2020house} perform indoor layout generation on these two datasets respectively. However, the number of rooms in each floor plan rarely exceeds thirteen, 
limiting the extensive ability of trained models from being applied to city-scale generation tasks. 

\textbf{3) City-level Datasets}: RoadNet~\citep{chu2019neural} contains real-world road network collected from OpenStreetMap (OSM) of 17 cities and  SpaceNet~\citep{vanetten2019spacenet} offers over 685,000 building footprints across 5 cities. These two datasets only involve 2D polyline or polygon data, which are limited to a certain extent considering  the complexity of 3D real-world applications. By contrast, RealCity3D is a city-level dataset that consists of not only 2D city layout information, but also building facade details. This enables more complex, large-scale city generation tasks. Recently, some 3D city-level datasets are released: HoliCity ~\citep{zhou2020holicity}, UrbanScene3D ~\citep{lin2022capturing}, SUM~\citep{Gao_2021}, and NYC3D~\citep{NYCdata_web}. However, most of these datasets are acquired from few cities, leading to a dearth of diversity in city layouts. In contrast, our RealCity3D dataset is compiled from eight major cities (can be easily extend to more cities), encompassing a broader spatial extent and featuring a wider range of city layouts than other city-level datasets. This inherent diversity renders our dataset particularly well-suited for tasks such as city layout generation, which demand a more varied data distribution.

\subsection{City Generation Methods}

Early works focus on reconstructing city models from LiDAR point clouds~\citep{bauchet2019city,huang2022city3d} or from images~\citep{zhao2022extracting,gui2021automated}. However, these reconstruction methods may not adequately satisfy the growing demands of various applications like gaming, urban planning, and autonomous driving, which require diverse city models. Generation methods present an opportunity to meet this rising demand by synthesizing a variety of city models. Existing research on city generation can be divided into two categories: non-data-driven and data-driven methods. Early studies mainly utilize procedural modeling to generate synthetic cities. In addition, with the development of deep learning, a few data-driven methods have been proposed to generate geometric data, which is related to city generation. 





\textbf{Procedural modeling}, such as L-systems, creates geometric structures based on handcrafted shape grammar ~\citep{merrell2010computer,vanegas2012procedural,yang2013urban,demir2014proceduralization}, a set of Euclidean shape transformation rules. In addition, the ESRI CityEngine, popular with the urban planning community, is a commercial generation engine that applies shape grammar to generate large-scale city layouts. However, these procedural generation methods require experts to manually adjust the rules and parameters. To automatically learn these rules, inverse procedural modeling uses deep neural networks to extract shape grammar from existing 2D and 3D datasets. ~\citep{vanegas2012inverse,ritchie2016neurally,nishida2016interactive,guo2020inverse}.


\textbf{Data-driven generation methods} have gained popularity in recent years as it enables the generation of complex geometric structures (vertices/lines/surfaces) with minimal human input. According to the aforementioned datasets, existing methods related to city generation can be categorized into three levels: (1) single object generation \citep{nash2020polygen,achlioptas2018learning,nauata2020housegan}, (2) indoor layout generation~\citep{nauata2020house, hu2020graph2plan} \citep{liu2018floornet} , and (3) city layout generation \citep{chu2019neural,zhang2020conv}. However, methods for single object generation neglect the geometric relationships between objects, and methods for indoor layout generation are only suitable for small-scale indoor generation tasks, thus neither of these two types of methods are able to directly apply to city generation. In addition, existing methods for city layout generation, including Neural Turtle Graphics (NTG)~\citep{chu2019neural}, RoadNet~\citep{chu2019neural} and  SpaceNet~\citep{vanetten2019spacenet}, mainly focus on the 2D city layout generation, without considering the 3D geometric characteristics of real cities. Recently, BlockPlanner~\citep{xu2021blockplanner}, represents city blocks via graphs and leverages Graph VAE to synthesize novel layouts.

\subsection{Tree-structured Neural Networks}

Tree-structured neural networks have long been explored for natural language processing tasks, such as program generation~\citep{chen2018tree}, sentence parsing~\citep{goller1996learning,socher2011parsing}, and representation learning~\citep{tai2015improved,li2015tree}. More recently, \citep{roy2020tree} proposed Tree-CNN, a model to grow neural networks during incremental learning. On the geometric data, researchers have explored tree networks for 3D point cloud modeling. \citep{klokov2017escape} proposed KD-Net that encodes point cloud features hierarchically with a KD-tree, and used it for classification and segmentation. \citep{gadelha2018multiresolution} extended it to a multi-resolution tree networks for more efficient point cloud processing. As for generation task, ~\citep{li2017grass} and ~\citep{mo2019structurenet} further extend the application of tree-structure network to geometric shapes generation. Similarly,  ~\citep{sharma2018csgnet} used a recurrent networks to parse the 2D/3D shapes into tree-structure executable programs. Moreover, \citep{li2019grains} proposes a recursive auto-encoder to generate indoor scenes based on the hierarchical structure. However, \ourname~focuses on much larger set of outdoor city layout data instead of a single 3D shapes or limited region of indoor scenes. 
\section{The RealCity3D Dataset}\label{sec:dataset}

\subsection{Data Standard and Collection}\label{Data Standard and Collection}

\begin{figure*}
    \centering
      \begin{minipage}[c]{1\textwidth} 
    \centering 
    \includegraphics[width=0.8\textwidth]{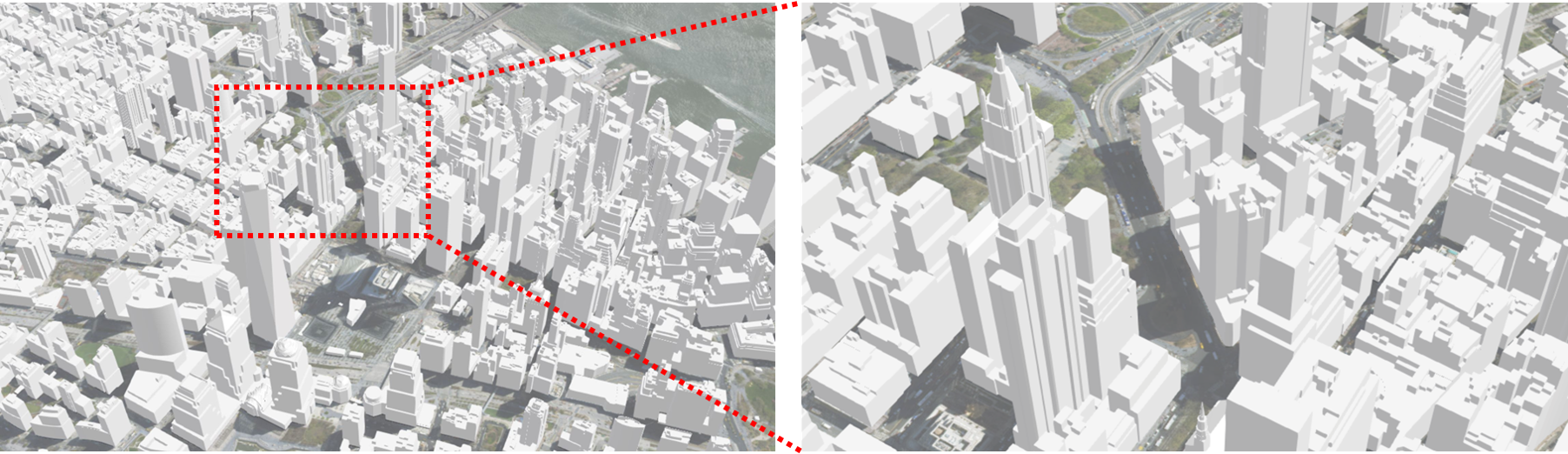}
    \caption{
     Detailed views of 3D building shapes in New York
    }
    \label{fig2_nyc_details} 
  \end{minipage}%
  
    \begin{minipage}[c]{1\textwidth} 
    \centering 
    \includegraphics[width=0.8\textwidth]{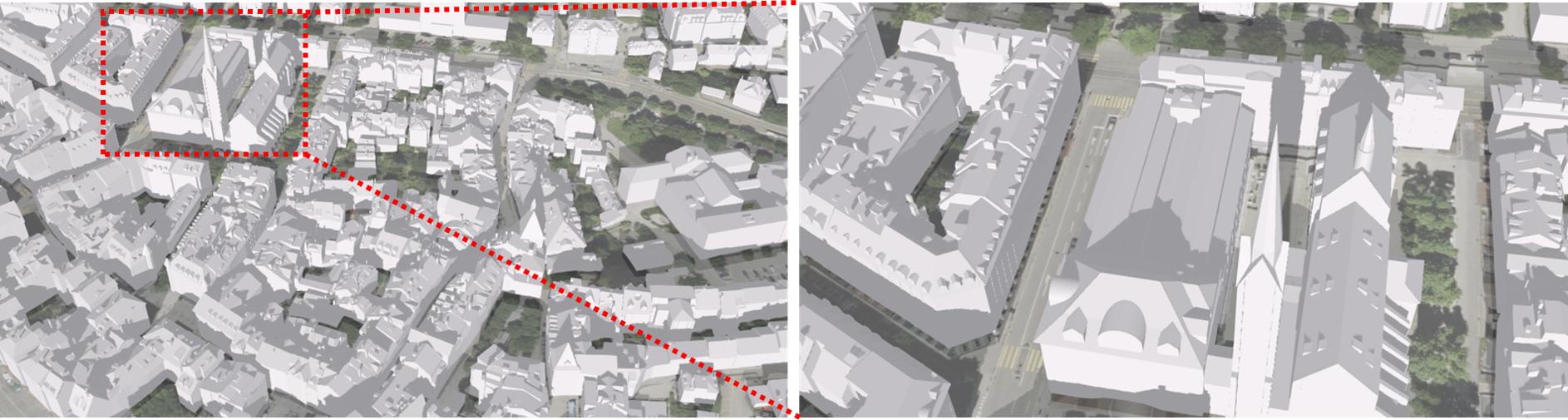}
    \caption{
     Detailed views of 3D building shapes in Zurich
    }
\label{fig2_zurich_details}
    \end{minipage} 
\end{figure*}

In our dataset, 3D building objects are extracted from 3D city models in CityGML format, an XML-based format widely used by the Architecture, Engineering and Construction (AEC) community for efficient storage of city-scale data. CityGML extends XML by adding sets of primitives, including topology, features, geometry, and city-specific constraints. Examples of 3D object classes in CityGML consist of buildings, tunnels, and bridges. CityGML has a hierarchical model complexity system to mark the complexity of each object class from LoD1 (Levels of Detail) to LoD4, as shown in Figure~\ref{fig2_citygml}. 


We collected CityGML data of New York, Zurich, Tokyo, Berlin, Boston, Vienna, Brussels, and Estonia, and we use data of New York(Figure~\ref{fig2_nyc_details}) and Zurich (Figure~\ref{fig2_zurich_details}) as examples to show how we process the date in this section. Note that these data are obtained from real-world using geomatics and surveying techniques. Since CityGML data from publicly available 3D geospatial datasets contain building models mostly with LoD2 complexity, we store all building objects in our dataset in LoD2 format. Besides, the CityGML data quality of different cities in our datasets varies considerably, presenting technical difficulties for scalable data processing. For example, only 76\% of Zurich buildings have valid CityGML surfaces, and the other 24\% have non-planar duplicated surfaces which violate the CityGML format standard. 
Our data processing pipeline, described in Figure~\ref{fig3_flow_processing}, can be scaled across cities where CityGML data are available.

\subsection{Dataset Processing}

\textbf{Parsing CityGML Files.} From each CityGML file, we extract building polygons and their semantic information as dictionaries. Only exterior components (i.e. polygon surfaces) are conserved. Recurring vertices are removed to ensure triangulation can be performed without error. Building surfaces are categorized as “GroundSurface”, “RoofSurface”, and “WallSurface” based on its CityGML semantic information. Semantically labelled polygons are output as .obj files for further transformation.

\begin{figure*}[h]
  \begin{center}
    \includegraphics[width=1\textwidth]{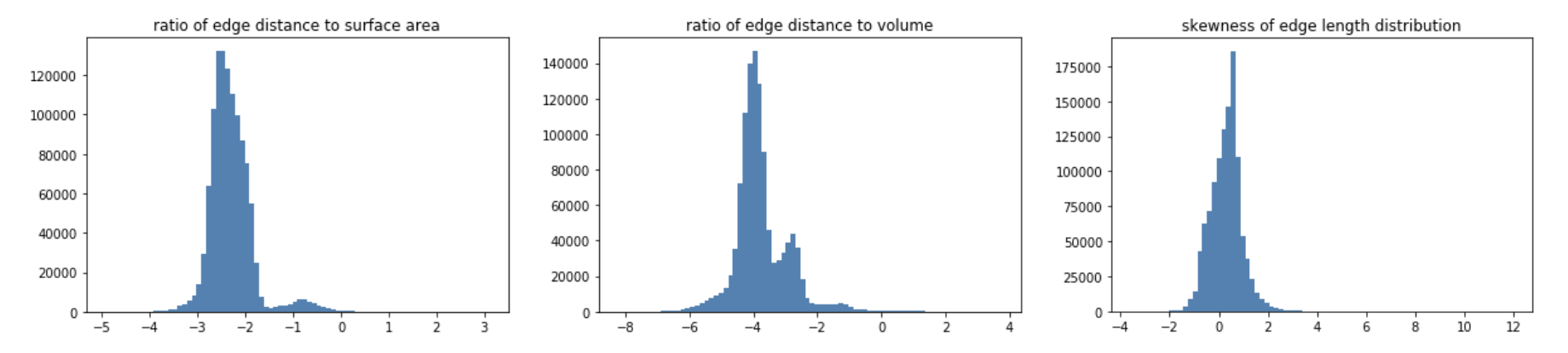}
  \end{center}
  \caption{Histograms of building shape complexity indicators.}
  \label{histograms_shape_complexity}
\end{figure*}

\textbf{Polygon Triangulation.} Taking a polygon mesh of a building as input, we apply polygon triangulation to decompose a polygon area into a maximal set of non-intersecting triangles on a continuous surface. \textit{n} vertices of a polygon are decomposed to \textit{n-2} triangles by triangulation. 

\textbf{Point Sampling.} We acquire 3D point clouds from each triangular mesh using Poisson disk sampling, a sequential and bias-free process for selecting points in each triangular subdomain. Poisson disk sampling has been used to achieve approximately uniform distance between adjacent points, yielding good visual resolution for rendering 3D buildings~\citep{yanai2017poisson}. By uniformly sampling these points on a continuous mesh surface, we reduce the amount of noise outliers that may come with conventional LiDAR scans of city buildings. 

\textbf{Voxelization.} To provide a greater geometric structure, we organize the 3D point cloud into a discrete voxel representation. Voxelization is a common method for downsampling and facilitating rapid retrieval of large-scale point cloud data~\citep{xu2021voxel}, which is essential in real-world urban planning applications. We employ the open-source binvox program which adopts a variation of the ray-stabbing method described in~\citep{fakir2003ray} to efficiently rasterize the point cloud into a 3D voxel grid. The ray-stabbing method classifies voxels as either an interior or exterior voxel by imagining a ray stabbing through the model. Voxels at the two extreme depth samples of the ray (i.e. when the ray first penetrates the model, and when it leaves the model) are classified as exteriors, while the rest are interiors. 


\subsection{Data Statistics}\label{data_stats}
1,135,403 building models were extracted and converted to polygon meshes, triangular meshes, point clouds, and voxels. 77\% of the dataset are non-cuboid shapes, while 4.9\% contain non-orthogonal components (e.g., sloping roofs). Building mesh statistics (Table \ref{mesh_statistics}) demonstrates the dataset's geometric variance. Figure \ref{histograms_shape_complexity} showcases its range of geometric complexity using three common indicators~\citep{abualdenien2020}: the ratio of edge lengths to surface area and volume, and the skewness of edge lengths.

\begin{table*}[hb]
\begin{center}
\caption{Building mesh statistics.}
\label{mesh_statistics}
\begin{tabular}{c c c c c c c c }
\hline
 & Vertices & Edges & Faces & Edge Length & Footprint Area & Surface Area & Volume\\
\hline
mean & 21.09 & 33.02 & 13.13 & 660.15 & 1,498.70 & 8,243.68 & 79,557.48\\
std & 31.17 & 49.98 & 18.03 & 1,212.98 & 5,019.27 & 20,803.83 & 660,395.00\\
min & 8.00 & 12.00 & 6.00 & 21.90 & 0.01 & 16.64 & 3.29\\
max & 5,148.00 & 7,722.00 & 2,768.00 & 254,437.40 & 1.17M & 3.12M & 215.98M\\
\hline
\end{tabular}
\end{center}
\end{table*}

\begin{table*}[h]
\begin{center}
\caption{Mean value of geometric features across regions.}
\label{geo_diversity}
\begin{tabular}{ c c c c c c c c }
\hline
 & Vertices & Edges & Faces & Edge Length & Footprint Area & Surface Area & Volume\\
\hline
Bronx & 23.54 & 36.52 & 14.26 & 838.04 & 2,111.66 & 11,121.01 & 118,532.07\\
Brooklyn & 17.69 & 27.62 & 11.42 & 624.18 & 1,532.30 & 8,302.48 & 66,686.16\\
Manhattan & 43.51 & 70.02 & 25.81 & 2,469.09 & 4,332.89 & 34,941.46 & 607,886.94\\
Queens & 16.65 & 26.02 & 10.71 & 515.44 & 1,265.43 & 6,231.68 & 47,713.10\\
Staten Island & 17.66 & 27.27 & 11.10 & 542.13 & 1,281.85 & 6,419.39 & 40,785.20\\
Zurich & 66.50 & 104.91 & 37.42 & 537.45 & 208.93 & 1,318.50 & 4,423.56\\
Berlin & 20.79 & 34.17 & 25.35 & 198.37 & 143.37 & 2272.59 & 4072.98\\
Boston & 34.75 & 93.31 & 67.06 & 1,804.61 & 123.21 & 8,520.70 & 105,656.00 \\
Brussels & 29.56 & 45.84 & 34.85 & 242.49 & 139.61 & 1,670.46 & 2,686.89 \\
Estonia & 18.80 & 28.75 & 12.04 & 173.93 & 43.64 & 602.76 & 1,397.41 \\
Tokyo & 74.49 & 117.49 & 44.12 & 682.62 & 278.06 & 2,374.83 & 1,4605.07 \\
Vienna & 66.63 & 96.51 & 68.21 & 445.87 & 167.96 & 3,061.24 & 5,345.03 \\
\hline
\end{tabular}
\end{center}
\end{table*}

Moreover, we obtain geometric statistics from various regions, including New York City's five boroughs, Zurich, Tokyo, Berlin, Boston, Vienna, Brussels, and Estonia. The results presented in Table~\ref{geo_diversity} reveal a distinction in geometric characteristics among the different areas, which poses significant challenges for the task of generating novel layouts in these regions. 
\section{\ourname}

Our goal is to achieve city generation based on a deep generative model. To this end, we first construct a binary tree from the spatial data which is explained in Section~\ref{sec:data_construction}, and then apply a customized tree-structured neural network for encoding, decoding and generation (Section~\ref{sec:network}). Figure~\ref{fig_overview} gives an overall architecture of our method by taking 2D parcels data as an example.

\subsection{Discovering Spatial Hierarchy in Data}
\label{sec:data_construction}
Let us first consider a set of spatial data $\left\{\mathcal{P}_{i} \right\}\mid_{i = 1, ..., N}$, where $\mathcal{P}_i$ represents a single object instance in the set and $N$ is the number of objects. In this paper,  we consider each building simplified by 3D cuboids to be an object instance, and we define $\mathcal{P}=(x, y, l, w, h, a)\in\mathbb{R}^6$, where $x$ and $y$ denote the center coordinates of a cuboid, and $l$, $w$, $h$ and $a$ denote the length, width, height and orientation angle of the cuboid. In order to preserve the spatial relationship among buildings, we organize the data hierarchically in a binary tree $\mathcal{T}$. To this end, we first regard all the original objects as leaf nodes and then generate $N-1$ intermediate parent nodes by merging the two closest leaf nodes. Finally, we can construct a binary tree until the generated parent node becomes the root node.


Specifically, we first define a set of distance metrics to measure the distance between two leaf nodes (i.e. building instances), which is utilized to search for a pair of leaf nodes to be merged. The formulations are defined as:
\begin{equation}
\begin{aligned}
     \quad{}D_{\text{center}}(i,j) = &\sqrt{(x_i - x_j)^2 + (y_i - y_j)^2} \\
     \quad D_{\text{area}}(i,j) = &\left|l_i  w_i - l_j  w_j\right| \\
     \quad{}D_{\text{shape}}(i,j) = &\left|(l_i / w_i) - (l_j / w_j)\right| \\
     \quad D_{\text{angle}}(i,j) = &\left|(a_i + a_j)/2 - a_{\texttt{MBR}(i,j)}\right| \\
     \quad{}D_{\text{merge}}(i,j) = &\left|l_i  w_i + l_j  w_j - l_{\texttt{MBR}(i,j)}  w_{\texttt{MBR}(i,j)}\right|
\end{aligned}
\end{equation}
where $D_{\text{center}}$ measures the Euclidean distance between the center points of two nodes; $D_{\text{area}}$ and $D_{\text{shape}}$ measures the difference between the area of two nodes and the aspect ratio of two nodes, respectively; $D_{\text{angle}}$ 
measures the difference in orientation between the mean of the two nodes and their minimum bounding rectangle (MBR), and $D_{\text{merge}}$ measures the difference in area between the sum of the two nodes and their minimum bounding rectangle (MBR).  Based on these distance metrics, we further introduce a spatial-geometric distance (SGD) metric to measure the similarity of building nodes in space and geometry, which can be formulated as: 
\begin{equation}
\begin{aligned}
D(i,j) = &\lambda_1 D_{\text{center}}(i,j) + \lambda_2 D_{\text{area}}(i,j) + \lambda_3  D_{\text{shape}}(i,j)\\ + 
     &\lambda_4 D_{\text{angle}}(i,j) + \lambda_5 D_{\text{merge}}(i,j) 
\end{aligned}
\end{equation}
where $\lambda_i$ represents the weight of each distance. And the value of $\lambda_i$ is determined empirically that will be discussed in Section~\ref{sec:hyper}.

Next, we recursively merge the two leaf nodes that are closest under the SGD metric to generate their parent node based on hierarchical clustering~\citep{johnson1967hierarchical}. To be more specific, for any produced intermediate node, its parameters $x$ and $y$ are obtained as the mean value of corresponding children nodes, and $l, w, h$ and $a$ are defined as the minimum bounding rectangle (MBR) of its children nodes. In addition, to make the model training faster and more stable, we calculate the relative parameters of child nodes with respect to the corresponding parent nodes after generating the parent nodes. This normalization operation naturally exploits the property of the tree structure and achieves better generative performance (discussed in Section~\ref{app:rela}).
We formulate this process as:
\begin{equation}
\begin{aligned}
     &x^{'}_{c} = \frac{x_{c}- x_{f}}{l_{p}},\quad y^{'}_{c} = \frac{y_{c}- y_{p}}{w_{p}},\\
     &l^{'}_{c} = \frac{l_{c}}{l_{p}},\quad w^{'}_{c} = \frac{w_{c}}{w_{p}}, \\
     &h^{'}_{c} = \frac{h_{c}}{h_{p}},\quad a^{'}_{c} = a_{c}- a_{p}. 
\end{aligned}
\end{equation}
where subscript $c$ means children node and $p$ means parent node.

\begin{figure}[h]
  \begin{center}
    \includegraphics[width=0.4\textwidth]{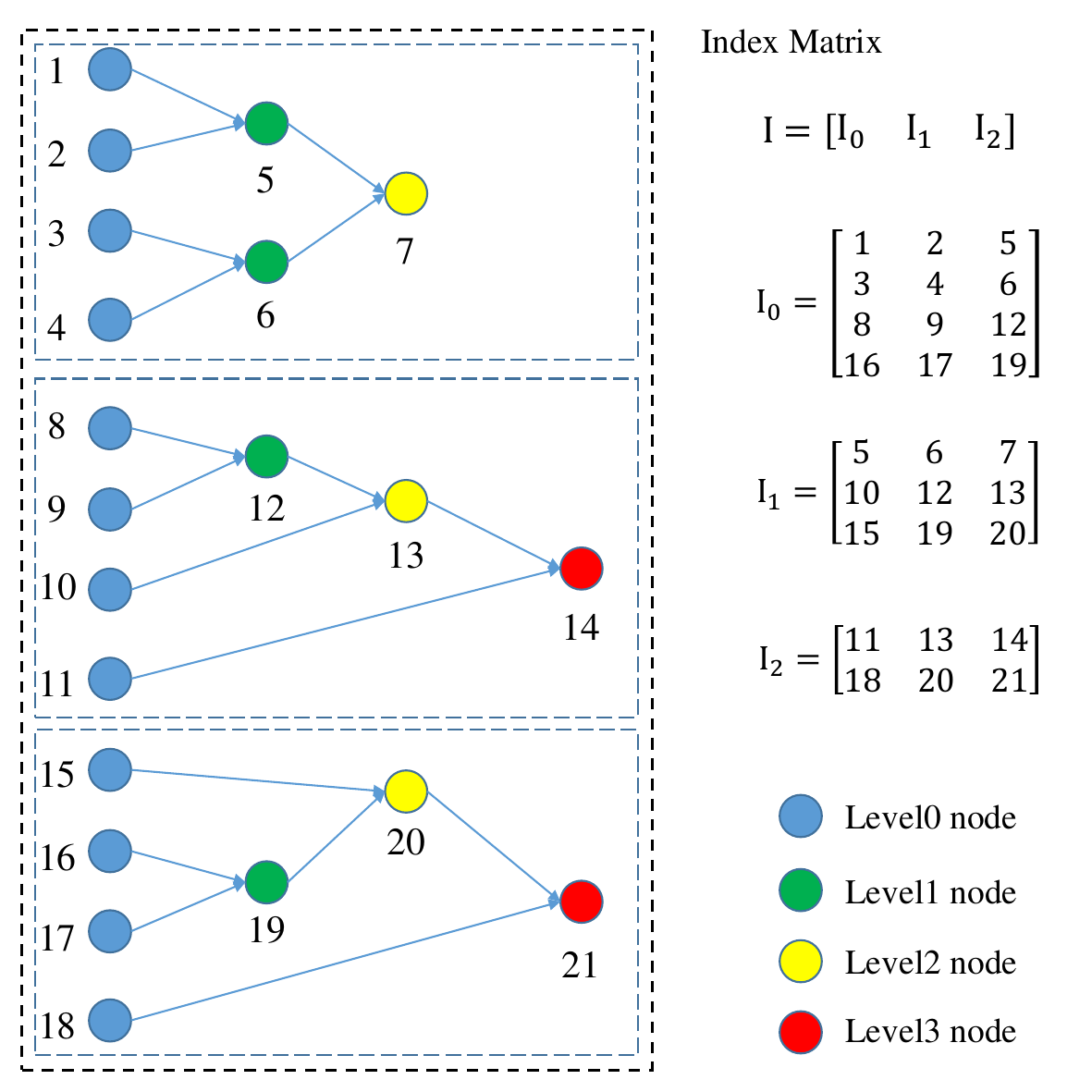}
  \end{center}
  \vspace{-10pt}
  \caption{An example of our data structure. The blue nodes at level 0 store raw objects and we build a hierarchical clustering tree according to their pairwise distances. For efficient mini-batch training, we design an index matrix for each level storing the indexes of children and parents. \label{fig3}}
\end{figure}

We present a simple example to explain our data structure. As shown in Figure~\ref{fig3}, we construct three trees based on the process described above, and each tree is outlined by a solid blue box in the figure. It can be found that the structure of the three trees is different although they are all built from the same number of leaf nodes, which makes neural network only capable of taking one tree as input during the training process. To this end, we present an index matrix $I$ to store the merging rules of different trees, where the first two columns denote the indices of children nodes and the last column is the index of the parent node. Therefore, we enable the network to process multiple trees with different structures at one time with the help of the index matrix, allowing the network to be trained and tested more efficiently.




\begin{figure*}[h] 
\centering
\includegraphics[width=16cm]{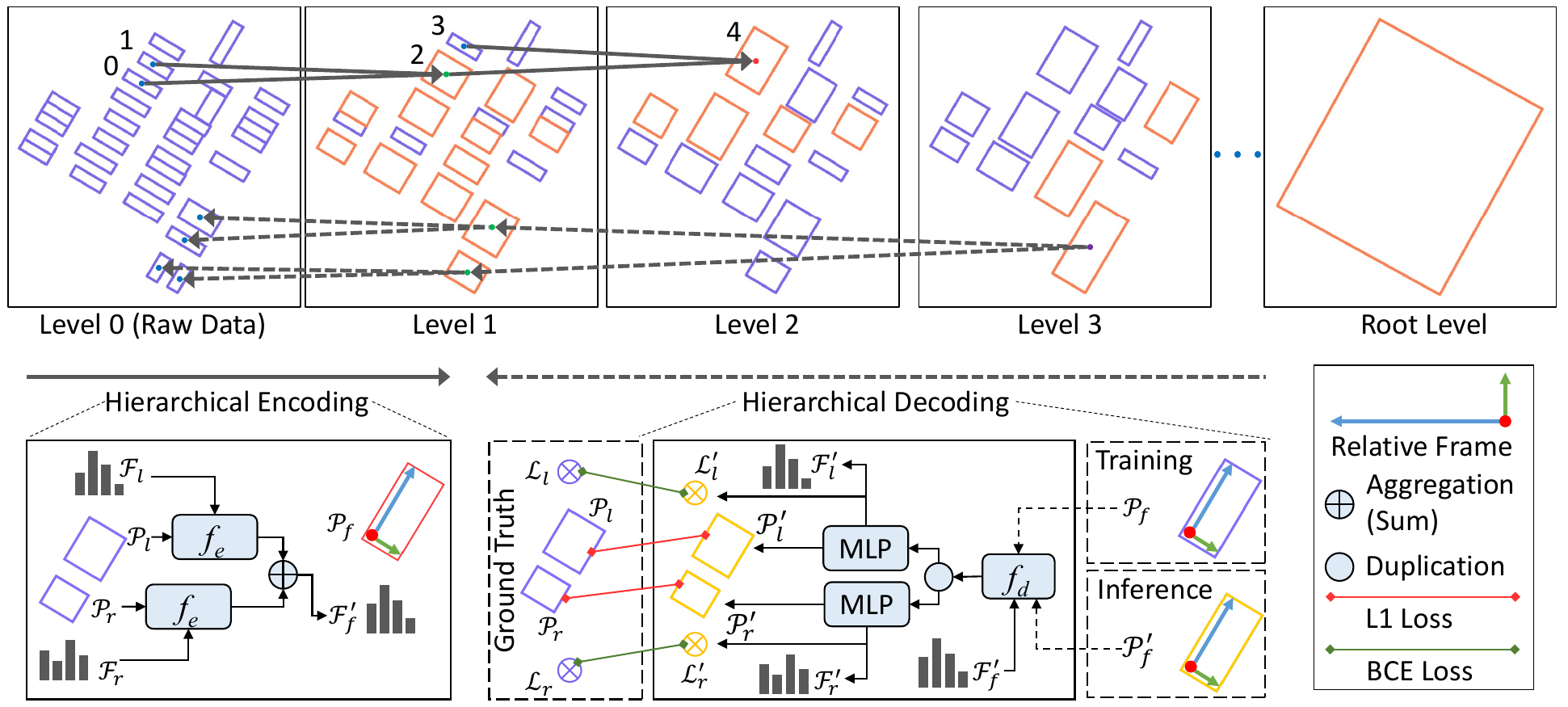}
\caption{Illustration of the \ourname~model on 2D parcels data example. The top row in the figure displays an example of pre-calculated tree structure from raw data level to root level. The orange boxes at the each level represent new boxes obtained by merging children boxes form the last level (for example, box 2 in level 1 is generated based on box 0 and 1 from level 0). The bottom row presents our encoding and decoding modules. Based on the tree structure, we first hierarchically encode children nodes to acquire the features of their father nodes until getting the root node features. Then starting from the root level, we hierarchically decode father nodes to children nodes and finally obtain the parameter of raw nodes. 
}
\label{fig_overview}
\end{figure*}

\subsection{Tree-shaped Auto-encoder Network}
\label{sec:network}

After hierarchically organizing the building object nodes into a binary tree, we naturally propose an auto-encoding tree neural network, namely AETree, to learn the latent representation of the tree by iteratively encoding nodes from leaf nodes to root nodes, and to reconstruct the original leaf nodes by decoding nodes from the root nodes. Then, based on the learned representation, we train a Gaussian Mixture Model (GMM) to estimate the probabilistic distribution of the representation. Finally, we generate a set of trees by decoding the new representations randomly sampled from the learned GMM.




\paragraph{Encoding.} Given a binary tree $\mathcal{T}$, $\mathcal{P}_x\in\mathbb{R}^6$ and $\mathcal{F}_x\in\mathbb{R}^C$  represent the geometric parameters and features of node x, respectively.
We first utilize an encoding function $f_e$ to recursively acquire the features of parent node $\mathcal{F}'_{f}$ based on the $\mathcal{P}$ and $\mathcal{F}$ of children node. In order to make the process of generation more flexible, we adopt a symmetric functions $\textit{g}$ to aggregate the information of left and right children nodes. The formulation can be expressed as follows:
\begin{equation}
\mathcal{F}'_{f} = \textit{g}(f_e([\mathcal{P}_{l},\mathcal{F}_{l}]), f_e([\mathcal{P}_{r},\mathcal{F}_{r}])),
\end{equation}
where $l$ and $r$ denote the left child node and right node. The function $f_e$ can be any feature extraction neural network, such as Multi-Layer Perceptron (MLP), Convolutional Neural Networks (CNNs) and Recurrent Neural Networks (RNNs). In this paper, to capture the long-dependency as the level of tree increases, we adopt the Long Short-Term Memory (LSTM) Cell as $f_e$ function. Each LSTM Cell consists of the hidden state $h$ as short term memory and the cell state $c$ as long term memory. Therefore, the features $\mathcal{F}$ of node can be regarded as a combination of $h$ and $c$. 

Specifically, we first initialize the feature $\mathcal{F}$ of the leaf node as zero vector. Then we split the feature $\mathcal{F}_l$ and $\mathcal{F}_r$ into $h_l$ and $c_l$, and $h_r$ and $c_r$, separately. Next, we fed the $\mathcal{P}_l$, $h_l$ and $c_l$ into LSTM Cell to obtain the updated $h^{'}_l$ and $c^{'}_l$ of left child node. Similarly, we can obtain the updated $h^{'}_r$ and $c^{'}_r$ of right child node. Note that the LSTM Cells of the left and right node are share-weighting. Finally, we utilize + operators as symmetric functions to calculate the $h^{'}_f$ and $c^{'}_f$ of parent node, which are concatenated as the $\mathcal{F}'_{f}$. The whole process can be formulated as:
\begin{equation}\label{eq:Encoding}
\centering
\begin{aligned}
     [h_l, c_l] = \mathcal{F}_{l}, &\quad [h_r, c_r] = \mathcal{F}_{r} \\
     h^{'}_l,\; c^{'}_l &= LstmCell(\mathcal{P}_l, (h_l, c_l)) \\
     h^{'}_r,\; c^{'}_r &= LstmCell(\mathcal{P}_r, (h_r, c_r)) \\
     h_f = h^{'}_l + h^{'}_r, &\quad c_f = c^{'}_l + c^{'}_r \\
     \mathcal{F}'_{f} &= [h_f, c_f]
\end{aligned}
\end{equation}
\paragraph{Decoding.} Starting from the leaf nodes, we are able to obtain the features of all non-leaf nodes by following the above encoding process. We assume the learned features of the root node to be $\mathcal{F}_{root}$. And the goal of the decoding process is to reconstruct the geometric parameters of leaf nodes, which is accomplished by hierarchically decoding the features of parent nodes starting with the root features $\mathcal{F}_{root}$. To be more specific, we utilize a decoding function $f_d$ to generate the geometric parameters $\mathcal{P}'$ and features $\mathcal{F}'$ of children nodes by taking the geometric parameters $\mathcal{P}_{f}$ and features $\mathcal{F}_{f}'$ of parent nodes as input. Note that we employ the teacher forcing algorithm~\citep{williams1989learning} to train the network, which makes the model's training faster and more efficient, and thus we input the ground-truth (pre-calculated) parameters $\mathcal{P}_f$ of parent nodes instead of the generated parameters $\mathcal{P}'_f$ by decoding the grandparent nodes of the former layer. In addition, we involve an indicator $\mathcal{L}'$ to denote whether the current node is a leaf node, i.e. to determine whether the current node should be further decoded or not. We formulate the decoding process as follows:
\begin{equation}
[\mathcal{P}'_{l},\mathcal{F}'_{l},\mathcal{L}'_{l}, \mathcal{P}'_{r}, \mathcal{F}'_{r}, \mathcal{L}'_{r}]= f_d([\mathcal{P}_{f},\mathcal{F}'_{f}])
\label{eq:Decoding}
\end{equation}
where $l$ and $r$ denote the left child node and right node. Similar to encoding function $f_e$, we employ LSTM Cell as the decoding function $f_d$. And the \textcolor{blue}{Eq.}~\ref{eq:Decoding} can be rewritten as follows:
\begin{equation}\label{eq:Decoding_LSTMCell}
\centering
\begin{aligned}
     [h'_f, c'_f] &= \mathcal{F}'_{f}, \\
     h^{'}_{lr},\; c^{'}_{lr} = &LstmCell([\mathcal{P}_f, \mathcal{F}'_{f}], (h'_f, c'_f)), \\
     [h'_l, h'_r] = {h}'_{lr}, &\quad [c'_l, c'_r] = {c}'_{lr} \\
     [\mathcal{P}'_{l},\mathcal{L}'_{l}] = MLP(h'_l), &\quad [\mathcal{P}'_{r},\mathcal{L}'_{r}] = MLP(h'_r) \\
     \mathcal{F}'_{l} = [h'_l, c'_l], &\quad \mathcal{F}'_{r} = [h'_r, c'_r] \\
\end{aligned}
\end{equation}
We first split the feature $\mathcal{F}_{f}'$ of parent nodes into hidden state $h'_f$ and cell state $c'_f$. And then we pass the concatenation of $\mathcal{P}_f$ and $\mathcal{F}_{f}'$, $h'_f$ and $c'_f$ to LSTM Cell and obtain $h^{'}_{lr}$ and $c^{'}_{lr}$ of left and right children nodes, which are further respectively split into $h^{'}_{l}$ and $h^{'}_{l}$, and $c^{'}_{l}$ and $c^{'}_{l}$. Next, we apply the MLP to the respective hidden states $h^{'}$ of the left and right children nodes to obtain the corresponding geometric parameters $\mathcal{P}'$ and indicator variable $\mathcal{L}'$. Finally, we concatenate the hidden state $h^{'}$ and cell state $c^{'}$ into the decoded feature $\mathcal{F}'$ of children nodes.

\begin{figure*}[ht] 
\centering
\includegraphics[width=0.9\textwidth]{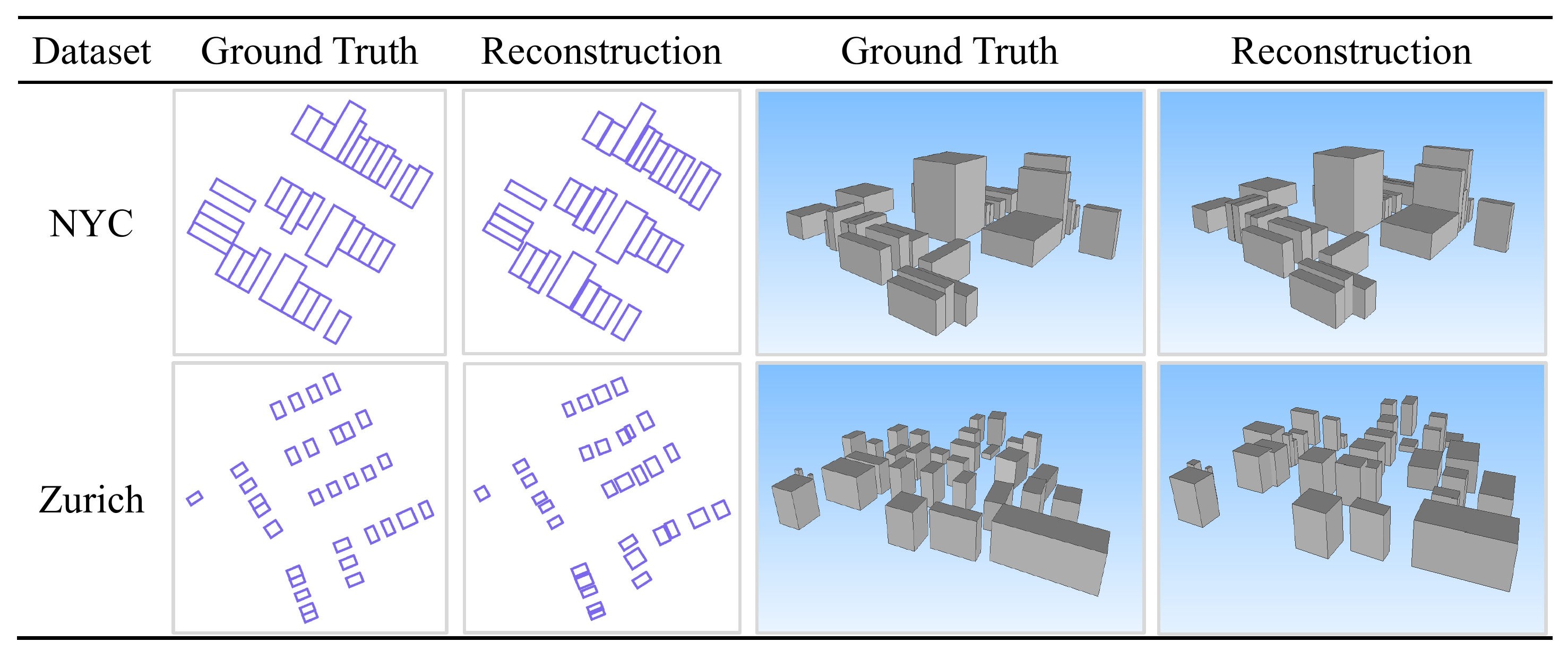}
\caption{2D and 3D building reconstruction results of \ourname~on the NYC and the Zurich dataset.}
\label{fig_rec_nyc_zurich}
\end{figure*}

\paragraph{Loss function.} We train our network by computing the L1 loss function between the ground-truth and predicted geographic parameters and the Binary Cross Entropy loss function between the ground-truth and predicted indicator variables that is used to identify whether the current node is a leaf node or not. Assume that the built binary tree has $M$ layers, the loss function can be formulated as follows:
\begin{equation}
Loss = \sum_m^M w_m (\| \mathcal{P}-\mathcal{P}'\| + \lambda [\mathcal{L}\log \mathcal{L}'+(1-\mathcal{L})\log (1-\mathcal{L}')],
\label{eq_inference}
\end{equation}
where $w_m$ denotes the weight coefficient of the current layer and $\lambda$ denotes the weight coefficient of the loss function of indicator variables. Note that the layers closer to the root node are given larger weights $w_m$, because the correct prediction of the previous layers in the decoding process is determinative for the generation of the following layers.

\paragraph{Inference and Generation.} After training our proposed model, when given the geometric parameters of a region and the features of the root node, we can infer the corresponding leaf nodes. Different from the training process, we input the decoded geometric parameters $\mathcal{P}'_{f}$ instead of the ground-truth geometric parameters $\mathcal{P}_{f}$ during the decoding phase, which is formulated as:
\begin{equation}
[\mathcal{P}'_{l},\mathcal{F}'_{l},\mathcal{L}'_{l}, \mathcal{P}'_{r}, \mathcal{F}'_{r}, \mathcal{L}'_{r}]= f_d([\mathcal{P}'_{f},\mathcal{F}'_{f}])
\label{eq:Inference}
\end{equation}
In order to enable our model to generate arbitrary and diverse leaf nodes, we first utilize a Gaussian Mixture Model (GMM) to estimate the probability distribution $P(\mathcal{F}_{root})$ of root node features. And then we sample a new feature $\mathcal{F}'_{root}$ from the estimated distribution $P(\mathcal{F}_{root})$. Finally, we hierarchically decode the sampled feature $\mathcal{F}'_{root}$ according to Eq. \ref{eq:Inference} to achieve generation.


\section{Experiments}\label{sec:exper}

\subsection{Dataset}

We collect CityGML models of the New York City (NYC) from ~\citep{nycwebsite}. Then, we extract 955,120 individual building models which are represented by polygon meshes through parsing the raw CityGML data. The semantic information of buildings is preserved in polygon mesh through adding a class label to each surface in our dataset. There are 3 categories for building surfaces: ground surface, roof surface, and wall surface. Among these, we extract the building footprint from ground surface and calculate the minimum bounding rectangle of each building footprint to get a single box. For each building, we perform the same process with its 32 neighbor buildings and consider the obtained 32 boxes as one set. Moreover, we obtain the 3D cuboid set by adding the height of building to 2D box set. In this paper, we select 45,487 cuboid sets, which are generated based on all buildings in Manhattan, as our dataset. Similarly, we obtain CityGML models of Zurich~\citep{zurichwebsite}, the largest city in Switzerland, and process these models to obtain 52,225 cuboid sets. For each dataset, 70\% of the data are chosen as training sets, 10\% as validation sets, and 20\% as test sets. Note that we conduct experiments both on 2D boxes and 3D cuboids datasets and we consider a 2D box as a simplified cuboid represented by $\mathcal(x, y, l, w, a)$, as described in Section \ref{sec:data_construction}.


\subsection{Evaluation Metrics}
To quantitatively evaluate areal spatial data generation, we first convert spatial data to point clouds (a cuboid is converted to the eight corner points) and then adopt three popular metrics~\citep{achlioptas2018learning}, including Jensen-Shannon Divergence (\textbf{JSD}), Coverage (\textbf{COV}), and Minimum Matching Distance (\textbf{MMD}). 

\textbf{JSD} measures the similarity of marginal distributions between the reference and generated sets. The distribution of data is calculated by counting the number of points in each discretized grid cell. \textbf{COV} measures the fraction of points in generated data that are matched to the corresponding closest neighbor points in the reference data. 
\textbf{MMD} measures the fidelity of the generated set with respect to the reference set by matching each generated point to the point in reference data with the minimum distance. 

For COV and MMD, we only select Chamfer Distance (CD) to compute the distance between two point clouds. We leave out Earth Mover's Distance (EMD) as it requires the number of instances of two sets to be equal, which is invalid for our generation. Moreover, compared to points, a unique aspect of the box data is their spatial extent. Therefore, we introduce another metric, Overlapping Area Ratio (\textbf{OAR}), which measures the ratio of overlapped area to the total area of all objects. The formulation of OAR is as follows:
\begin{equation}\label{eq:quaternion}
OAR(O) = \frac{\sum_{o \in O}{}A(o),\quad if (o \cap \hat{o}), \forall \hat{o} \in \left\{O-o\right\} }{\sum_{o' \in O}{}A(o')},
\end{equation}
where A($\cdot$) is the object $o$ area, $\cap$ indicates two overlapping objects, $O$ is the set of generated objects.

\begin{figure*}[t] 
\centering
\includegraphics[width=0.8\textwidth]{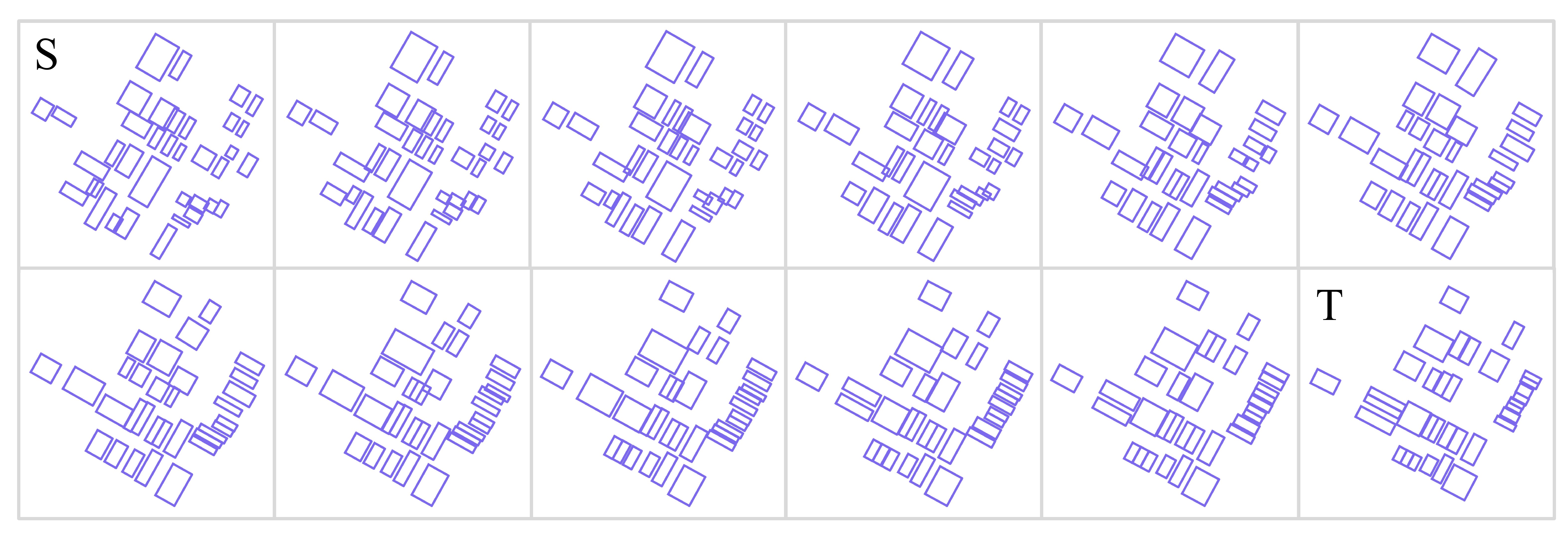}
\caption{Latent space interpolation between two box sets, S and T, using \ourname.}
\label{fig_interpolation}
\end{figure*}

\subsection{Implementation Details}
As stated in Section \ref{sec:data_construction}, a comprehensive metric which consists of five distance with different weights is defined for hierarchical clustering. After conducting a series of comparison experiments, the weights $\lambda_{1-5}$ are set to 5, 2, 0.1, 1, and 1 respectively. Meanwhile, the GMMs of 60 and 80 components with full covariance are selected as generators for the NYC and the Zurich dataset, respectively.

The \ourname~model is implemented based on the Pytorch framework. We employ the ADAM optimizer with a learning rate of 0.001 and decrease the learning rate by two times every 400 steps. The batch size is set to 50 in our experiments. For the JSD metric, the number of points is counted by discretizing the space into $28^3$ voxels.

\subsection{Results}
\subsubsection{Reconstruction.}
We first present the reconstruction results of our models. To show the effectiveness of different encoding and decoding functions, we employ the MLP and the LSTM Cell as basic function separately. For quantitatively evaluating the results, we transform spatial data to point clouds and use the CD, the EMD, and the OAR as evaluation metrics. Note that the reason why we can employ the EMD as the reconstruction metric here is that the number of the reconstructed and reference spatial data are equal. Table~\ref{table_comp_mlp_lstm} summarizes the comparison results of models with two base functions on the NYC and the Zurich Dataset. It can be seen that the LSTM Cell module performs better than the MLP as a base function. Therefore, the following results are all based on \ourname~with the LSTM Cell if no special explanation is provided. Moreover, we plot the reconstruction results in Figure~\ref{fig_rec_nyc_zurich}, which demonstrate the promising performance of our model.

\begin{table}[h]
\begin{center}
\caption{Quantitative comparisons of \ourname~with different base nets on the NYC and the Zurich dataset. $\uparrow$: the higher the better, $\downarrow$: the lower the better. The best values are highlighted in bold.}
\vspace{2mm}
\label{table_comp_mlp_lstm}
\scalebox{0.9}{\begin{tabular}{c c c c c}
\hline
Methods & Dataset & CD($\downarrow$) & 
EMD($\downarrow$)  &
OAR(\%, $\downarrow$) \\
\hline
\ourname~(MLP) & NYC & 0.0079 & 0.1206  & 5.81\\
\ourname~(LSTMCell) & NYC & \textbf{0.0019} & \textbf{0.0417} & \textbf{0.57}\\
\hline
\ourname~(MLP) & Zurich & 0.0077 & 0.1163 & 2.04\\
\ourname~(LSTMCell) & Zurich & \textbf{0.0027} & \textbf{0.0580}  & \textbf{0.14}\\
\hline
\end{tabular}}
\end{center}
\end{table}
\vspace{-10pt}
 

\subsubsection{Latent Space Interpolation.}
Given the latent codes of two samples, we are able to obtain the intermediate results by applying the decoder to the linear interpolation between these two latent codes. Therefore, we conduct an interpolation experiment on two randomly selected box sets to intuitively demonstrate the generative capability of the latent codes learned by~\ourname. Figure \ref{fig_interpolation} shows the reconstructed box sets from the interpolated latent vectors. Interestingly, we produce a gradually varied sequence of results from box set S to box set T, which demonstrates the smoothness of our latent space. Meanwhile, it can be found that our learned latent representations are generative instead of simply being able to memorize the training sets.

\begin{table*}[t]
\begin{center}
\caption{Quantitative comparisons of city layout generation performance of \ourname~with various data-driven baseline methods. The first four columns represent the results of models under four generation evaluation metrics and the last two columns measure the complexity of models. *: It is noted that LayoutTransformer achieves the best performance on metrics COV and MMD. However, as shown in Figure~\ref{fig_comparsion}, LayoutTransformer is not capable to generate visually high-quality layouts as \ourname, and it required much more computing resources than our model. A more detailed discussion is provided in Section~\ref{Generation}}. 

\label{table_comparison}
\begin{tabular}{c c c c c c c}
\hline
Methods & JSD($\downarrow$) & COV(\%, $\uparrow$) & MMD($\downarrow$)  & OAR(\%, $\downarrow$) & \#params & FLOPs/sample\\
\hline
SketchRNN-R2 & 0.0089 & 33.62 & 0.0050 & 1.83 &  2.19M & 243.13M\\
SketchRNN-R5 & 0.0101 & 28.76 & 0.0047 & 95.41 & 2.37M & 402.46M\\
PointNet-MLP & 0.0417 & 4.60 & 0.0219  & 87.47 & 1.84M & 3.67M\\
PointNet2-MLP & 0.0407 & 22.36 & 0.0086  & 56.39 & 2.03M & 31.55M\\
LayoutTransformer & 0.0041 & \textbf{68.91}* & \textbf{0.0037}*  &28.24 & 2.53M & 408.39M \\
LayoutVAE & 0.0061 & 41.56 & 0.0045 & 19.01 & 5.76M & 146.71M \\

\ourname & \textbf{0.0033} & 39.53 & 0.0044 & \textbf{1.66} & 2.91M & 31.86M\\

\hline
\end{tabular}
\end{center}
\end{table*}

\begin{figure}[t] 
\centering
\includegraphics[width=0.48\textwidth]{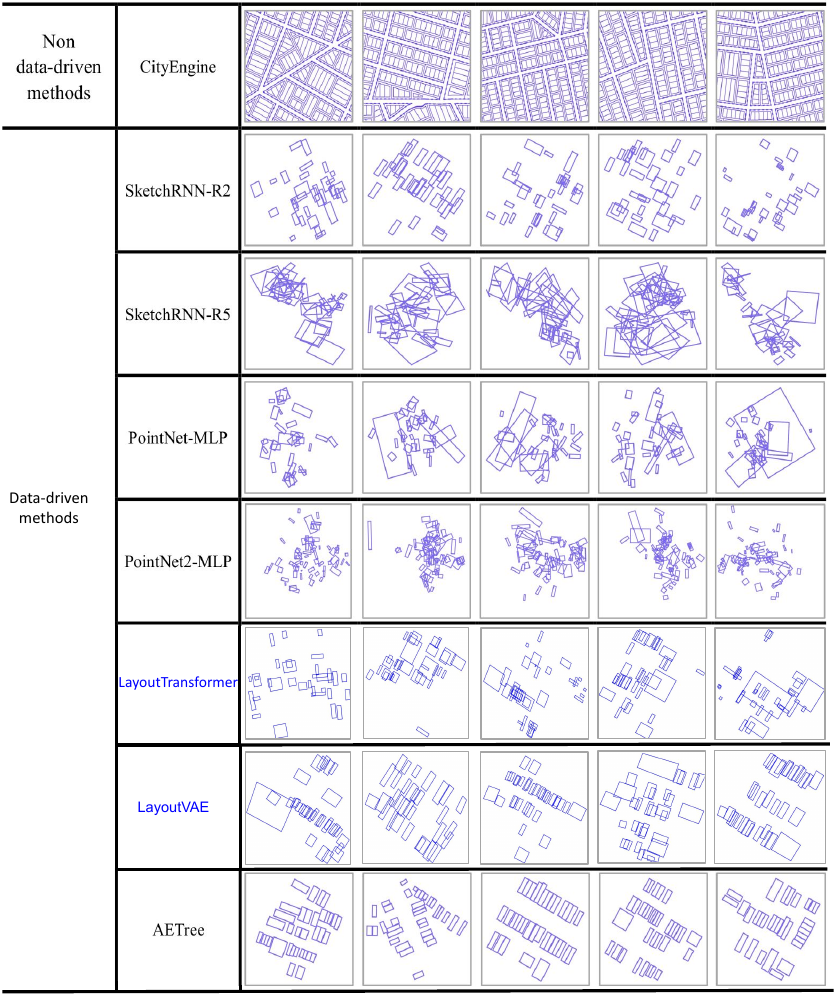}
\caption{City layout generation results of the models trained on NYC dataset.}
\label{fig_comparsion}
\end{figure}

\subsubsection{Generation.} \label{Generation}
After acquiring the generative latent representations, we fit Gaussian Mixture Models (GMMs) and sample new latent representations to generate new box sets or cuboid sets. Similarly, we first conduct a series of experiments on 2D building layout generation for intuitive comparison based on two types of baselines: procedural modeling through CityEngine and various data-driven methods.


\textbf{CityEngine}. ESRI CityEngine is a commercial software that uses a procedural modeling approach based on L-systems to create large-scale city models. By creating road networks and dividing the parcels into lots, it generates buildings on the allotments using predefined rules and parameters. The building footprint is generated using default rules with some manual adjustment of parameters.

\textbf{SketchRNN-R2}. SketchRNN is a generative model to generate sketch drawings~\citep{ha2017neural}. This model seems intuitively suitable to solve our problem. We convert each city layout data to a list of points as a sketch with x,y coordinates according to the input format of the SketchRNN. Specifically, for a batch with 32 building footprints, the converted sketch consists of 128 points by taking all corners. 

\textbf{SketchRNN-R5}. Based on the vanilla SketchRNN, we explore replacing the parameter of a city layout data (i.e. the center coordinates, length, width, height and orientation angle ) with x,y coordinates of a sketch. So we transform a batch of data to a sketch with 32 high dimension points, which incorporates 5 elements: ($\Delta x$, $\Delta y$, $\Delta l$, $\Delta w$, $\Delta a$). The first five elements are the offset parameters from the previous box. Different from~\citep{ha2017neural}, we use 1D to represent the binary state of the pen (at its end or not), since we assume that the pen draws 32 points in succession. 

\textbf{PointNet-MLP}. In addition, we benchmark a simple baseline model, which adopts PointNet~\citep{qi2017pointnet} as the encoder by regarding a city layout (corner points) as point clouds. By referring to the decoder of the SketchRNN, we employ MLP to decode the latent codes to represent for a probability distribution of points. Meanwhile, the loss function aims to maximize the log-likelihood of the generated probability distribution to explain the training data.

\textbf{PointNet2-MLP}. Moreover, we replace the PointNet with PointNet++~\citep{qi2017pointnet++} as model encoder while keeping the others same as PointNet-MLP.

\textbf{LayoutTransformer.} LayoutTransformer~\citep{gupta2021layouttransformer} is used to generate layouts in diverse natural as well as human designed data domains, which is a similar task to ours.  We organize the city layout data in ascending order based on the x-coordinate of each box's center for the ordered data requirement of this model. Furthermore, each box is represented by its four corners, which amounts to 8-dimensional point that can be directly utilized by this model. Since the number of boxes in each layout data is fixed as 32, there's also no need to add end token.

\textbf{LayoutVAE.} LayoutVAE~\citep{jyothi2019layoutvae} is a model used for generating scene layouts from textual description. Similar to LayoutTransformer, we use four vertices of the each box as inputs. In our experiments, the buildings are not categorized, so each building is labeled as 1.

Table~\ref{table_comparison} and Figure~\ref{fig_comparsion} present the quantitative and qualitative generation results of the city building layout, respectively. The results demonstrate that most data-driven baseline models do not perform well on our city layout generation dataset. It is noted that LayoutTransformer achieves better performance on two metrics, COV and MMD. However, according to Figure~\ref{fig_comparsion}, the quality of the generated layouts are still inferior to that of \ourname~, which can generate more regular and non-overlapped layouts. This indicates that the adapted metrics COV and MMD may not accurately assess the performance of this city layout generation task. Conversely, \ourname~demonstrates a significant performance advantage over LayoutTransformer on our proposed metric OAR. This underscores that our metric consistently provides a representative evaluation on the generation quality of city layouts.



\begin{figure*}[h] 
\centering
\includegraphics[width=0.8\textwidth]{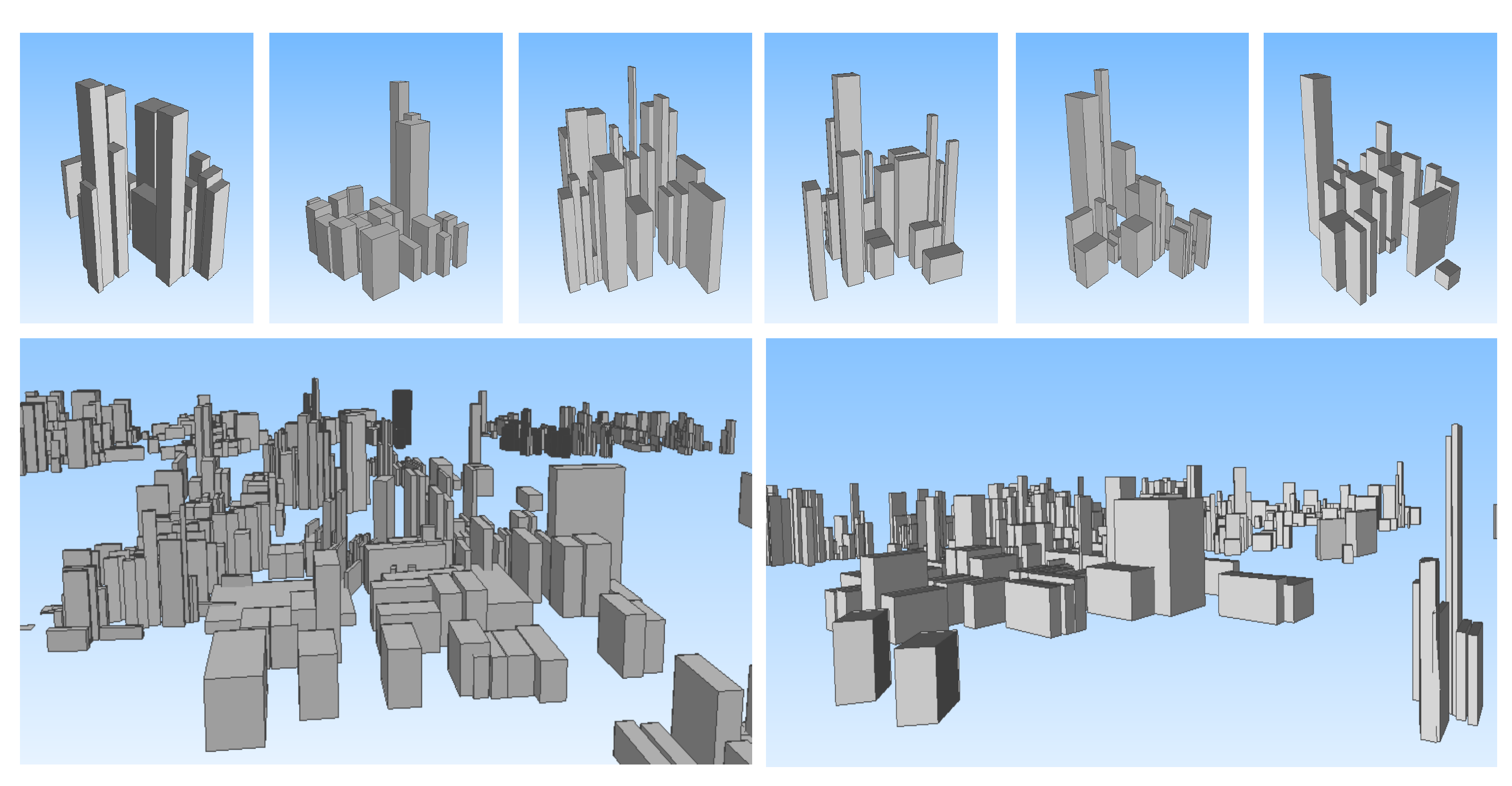}
\caption{3D city generation results of \ourname~trained on the NYC Dataset.
}
\label{append_fig_nyc3d_gen}
\end{figure*}

In addition, the learnable parameters and floating point operations (FLOPs) of each method are presented in the last two columns of Table \ref{table_comparison}. It can be found that the number of parameters of all methods is very close, but their FLOPs differ greatly. Though the PointNet-MLP model demonstrates the lowest complexity both in terms of the number of parameters and FLOPs, its generation results are unsatisfying on the four evaluation metrics. On the other hand, LayoutTransformer generates acceptable results, but it requires a much higher computational cost. By comparison, our model achieves the best generation performance with a relatively lower computation complexity. Furthermore, we show the 3D city generation results at small and large scales in the top and bottom rows of Figure~\ref{append_fig_nyc3d_gen}, respectively. 

\section{Discussion}

\begin{table}[h]
\begin{center}
\caption{The reconstruction comparisons of different distance metrics for hierarchical clustering.}
\label{table_ab_distance}
\resizebox{0.5\textwidth}{!}{%
\begin{tabular}{c c c c c c c c c }
\hline
$\lambda1$ & $\lambda2$ & $\lambda3$ & $\lambda4$ & $\lambda5$ & CD($\downarrow$) & EMD($\downarrow$)  & OAR(\%, $\downarrow$)\\
\hline
5 & 2 & 0.1 & 1 & 1 & \textbf{0.0019} & \textbf{0.0417}  & 0.57 \\
25 & - & - & - & - & 0.0021 & 0.0449 &  0.45 \\
- & 10 & - & - & - & 0.0019 & 0.0419 &  0.50 \\ 
- & - & 0.5 & - & - & 0.0022 & 0.0456 & 0.93 \\
- & - & - & 5 & - & 0.0020 & 0.0421 & \textbf{0.41} \\
- & - & - & - & 5 & 0.0042 & 0.0760 & 5.65  \\
\hline
\end{tabular}
}
\end{center}
\end{table}
\vspace{-10pt}
\subsection{Hyperparameter Analysis}\label{sec:hyper}



\textbf{Distance metrics}. As described in Section \ref{sec:data_construction}, we defined a comprehensive distance metric, a weighted sum of five distance metrics. The weight $\lambda$ of each distance is determined by comparison experiments, and the values of weight in our experiments are listed in the first row of Table \ref{table_ab_distance}. To explore the effect of each distance, we multiply 5 by the value of $\lambda_{i}$ respectively. As presented in Table \ref{table_ab_distance}, there is no big difference between these results except the change of $\lambda_{5}$, which means that the tree structure may be less effective if we put too much weight on minimizing $D_{\text{merge}}$.

\textbf{GMM parameters}.
To determine the optimal number and covariance type of Gaussian components for the GMM, we conduct a grid search using the JSD criterion.
As shown in Table \ref{table_ab_gmm}, the GMM of 60 components with full covariance obtains the optimal JSD value, which is thereby chosen for our generation experiments.
\begin{table*}[h]
\begin{center}
\caption{The generation results of GMMs with a varying number and covariance type of Gaussian components under JSD metrics. Each GMM is trained on the latent space learned by \ourname.}
\label{table_ab_gmm}
\resizebox{0.95\textwidth}{!}{%
\begin{tabular}{c c c c c c c c c c c}
\hline
\diagbox{cov. types}{\#components} & 10 & 20 & 30 & 40 & 50 & 60 & 70 & 80 &  90 & 100 \\
\hline
full & 0.0044& 0.0039& 0.0037& 0.0036& 0.0037& \textbf{0.0033}& 0.0034& 0.0034& 0.0035& 0.0034 \\
diag & 0.0176& 0.014 & 0.0177& 0.0118& 0.0127& 0.0171& 0.0105& 0.0118& 0.0133& 0.0255 \\
tied & 0.0057& 0.0042& 0.0045& 0.0040 & 0.0043& 0.0037& 0.0043& 0.004 & 0.0039& 0.0043 \\
spherical & 0.0057& 0.0060 & 0.0060 & 0.0059& 0.0059& 0.0057& 0.0063& 0.0061& 0.0061& 0.0062 \\
\hline
\end{tabular}%
}
\end{center}
\end{table*}

\begin{figure}[h] 
\centering
\includegraphics[width=0.48\textwidth]{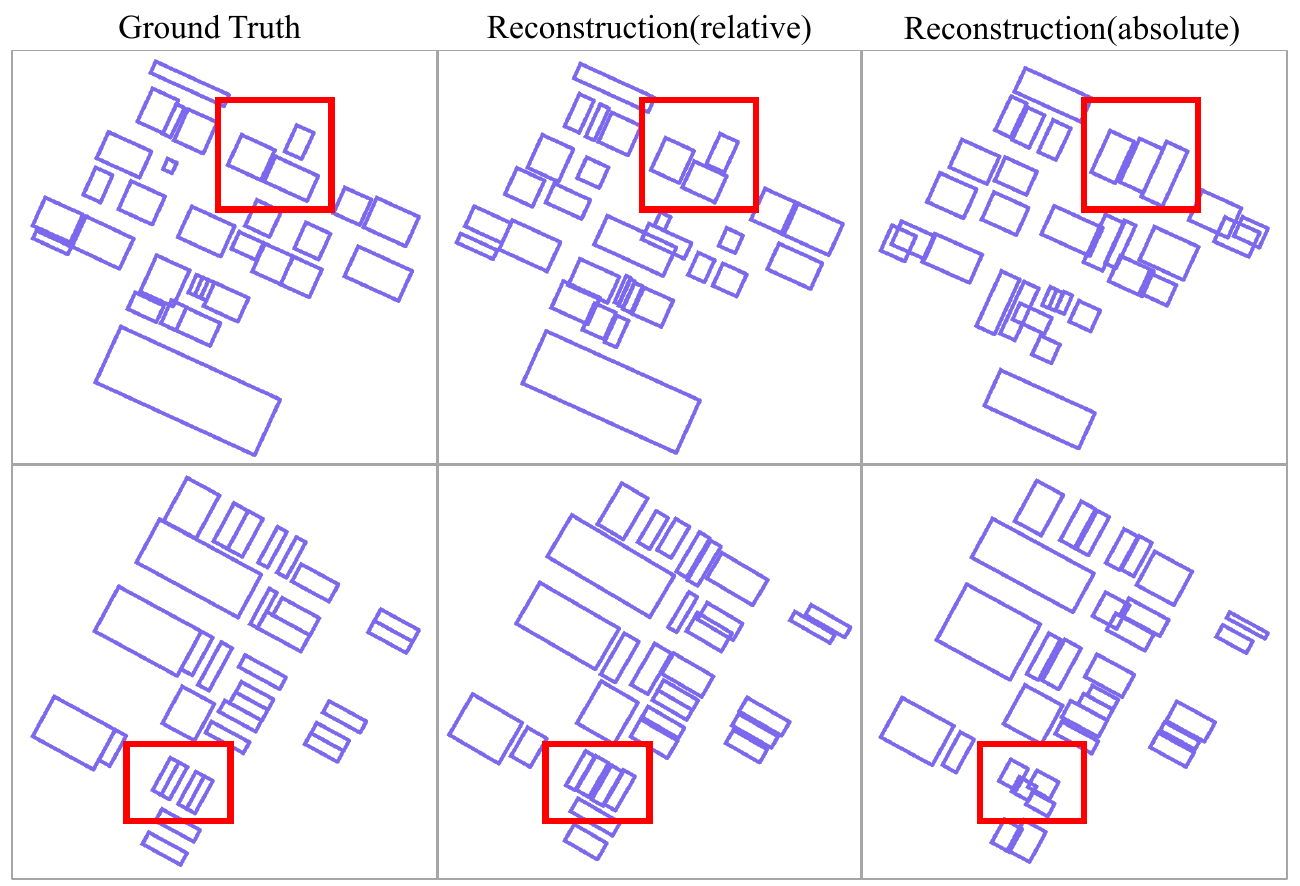}
\caption{Reconstruction results of models with the relative and absolute representation of spatial data. The red boxes highlight the most obvious difference between results.
}
\label{append_fig_dis_re_ab}
\end{figure}

\textbf{Discussion about relative representation.}\label{app:rela}
As described in Section \ref{sec:data_construction}, we preprocess all nodes' parameters relative to their father nodes. To verify the effectiveness of this relative representation, we conduct a comparison experiment between different ways of representation. In Figure \ref{append_fig_dis_re_ab}, we plot the reconstruction results of models with relative and absolute representations on the NYC dataset. It can be found that relative representation keeps the spatial relations among objects better than absolute representation, which is thus employed in our tree-based  structure data. 

\subsection{Latent Code Application}\label{sec:latent_ana_app}

Based on the latent codes encoded by \ourname, we further conduct two urban planning applications on the Manhattan region: classification of urban layout typologies and measurement of neighborhoods' urban layout composition.

\begin{figure}[t] 
\centering
\includegraphics[trim={160 0 140 0},clip,width=0.5\textwidth]{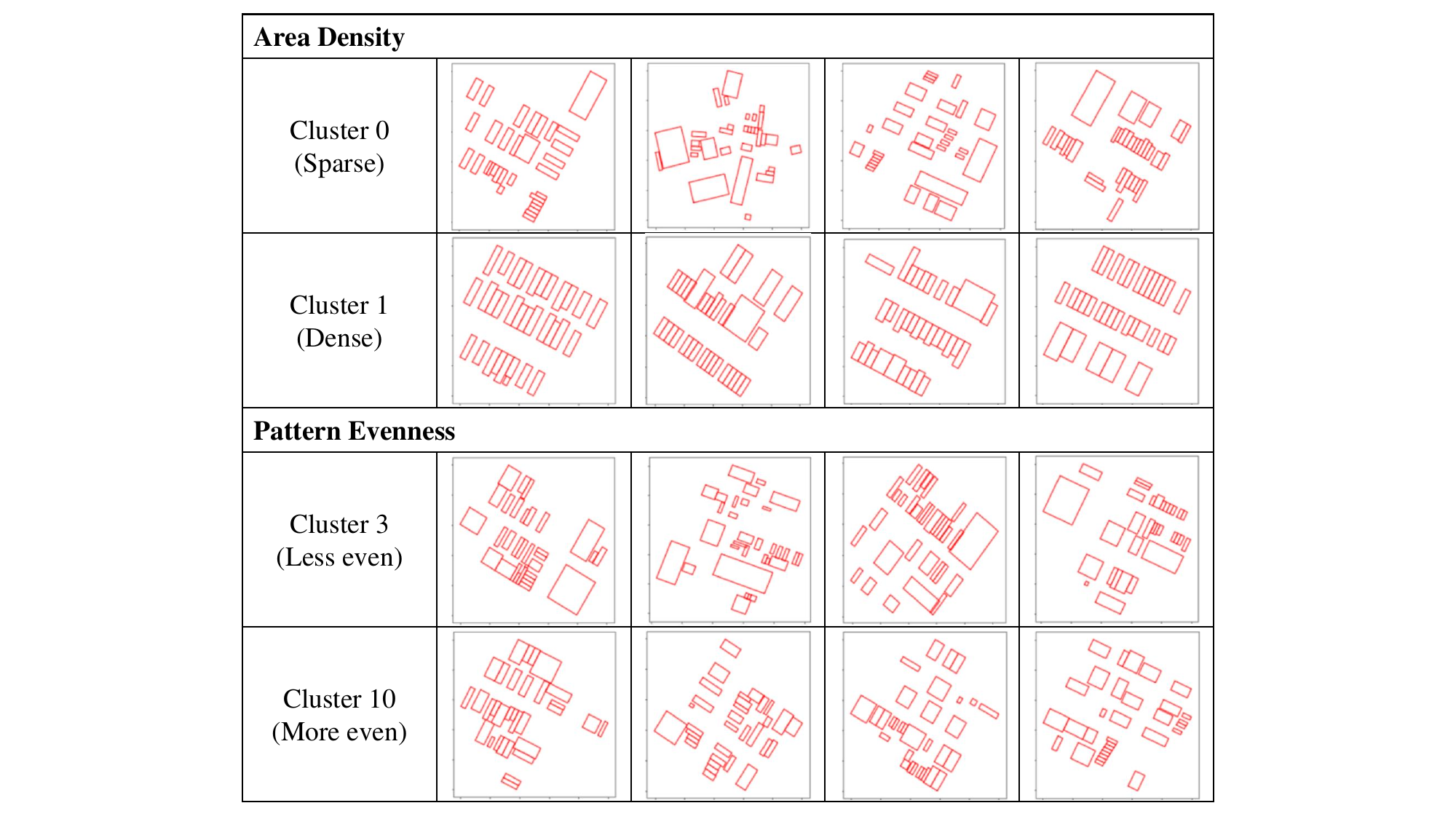}
\caption{Patterns of urban layout typologies in terms of geometric features.}
\label{cluster_visualize}
\end{figure}

\begin{table*}[t] 
\centering
\caption{Normalized mean values of geometric features of each cluster.}
\includegraphics[trim={150 340 150 10},clip,width=0.98\textwidth]{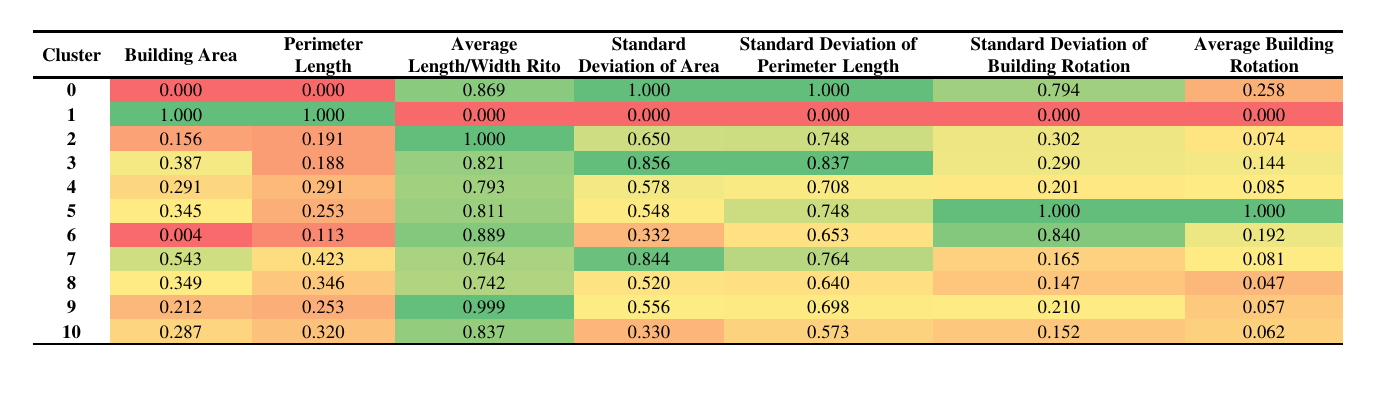}
\label{table_cluster}
\end{table*}

\textbf{Classification of urban layout typologies.} There is few standardised and data-driven way for classifying urban layouts, making it inefficient to objectively describe corresponding typologies. However, the latent representations learned by \ourname~can naturally serve as meaningful geometric features for categorizing the city layouts. Specifically, we firstly get the 512-dimension latent features of the 45,000 Manhattan's urban layouts using the trained \ourname's~encoder. Then, the principal component analysis (PCA) is applied to map the original latent features to 50-dimension latent codes. Based on these low-dimensional latent codes, we apply Gaussian Mixture Model clustering algorithm (GMM) to cluster the 45,000 urban layouts to 11 groups.

\begin{figure}[t]
\centering
    \includegraphics[trim={0 120 0 30},clip,width=0.5\textwidth]{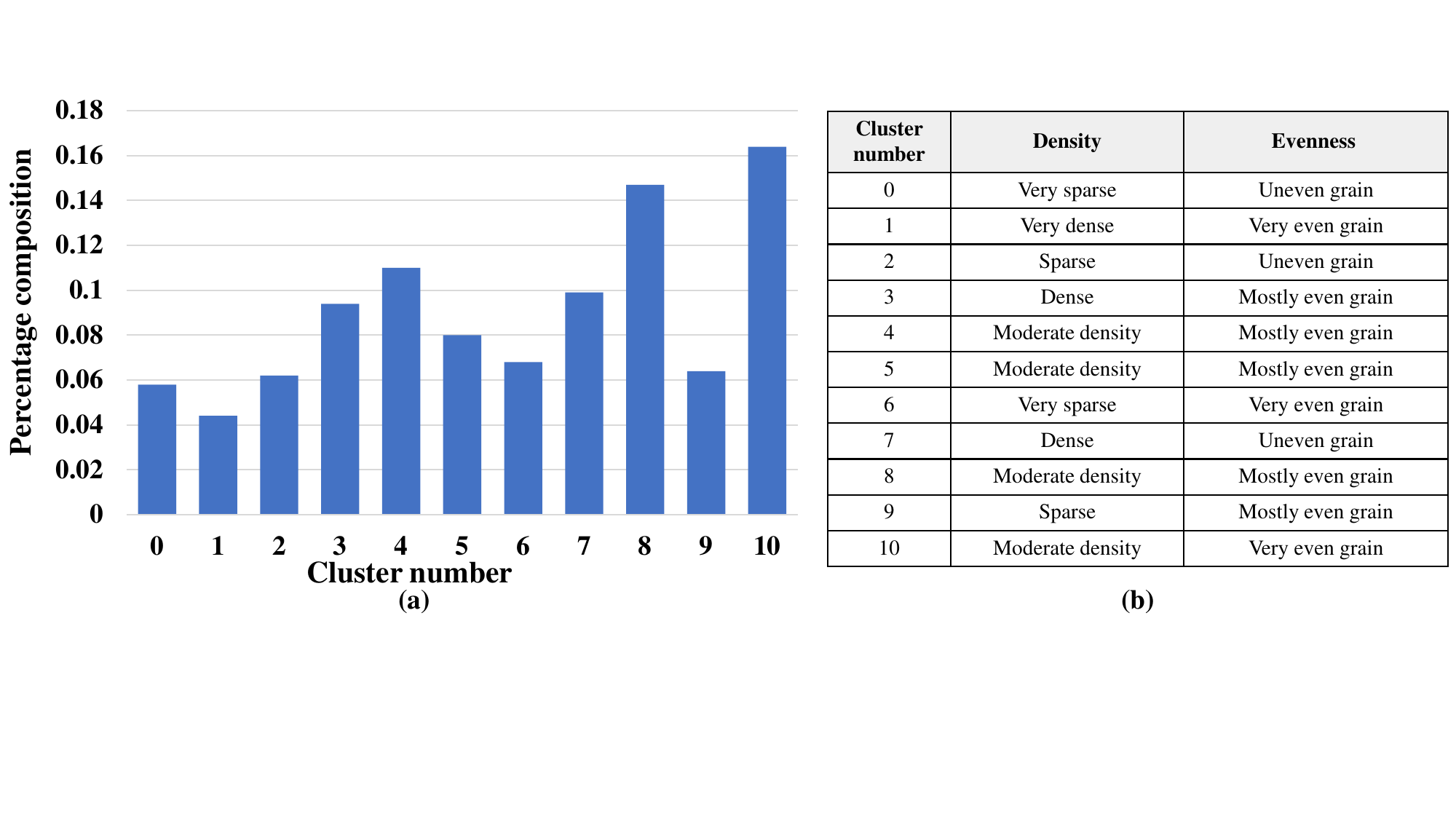}
  \caption{(a) Manhattan’s urban layout composition. (b) Detailed profiles of urban layout typologies of each cluster.\label{composition_figure}}
\end{figure}

Then, based on the latent space clusters identified by GMM, we further investigate if we can identify distinct clusters using several common-used geometric features in urban planning area such as: building area, perimeter length, average length/width ratio and building rotation, standard deviation (std) of area, perimeter length, and building rotation. From Table~\ref{table_cluster} (all values are scaled to the range of 0 to 1), we can see that some clusters have distinct values for geometric features. This indicates that the clusters were not randomly grouped, but were classified based on the aforementioned geometric features. 

Moreover, we identify two patterns of urban layout typologies: area density and pattern evenness, to further validate the effectiveness of the latent representation for classifying urban layout typologies. As shown in Figure~\ref{cluster_visualize}, Cluster 0, characterized by the lowest average building footprint area and length of building perimeter among all clusters, is composed of small buildings with ample spacing between them. In contrast, Cluster 1 exhibits the highest average building footprint area and length of building perimeter, indicating a qualitatively more dense distribution of buildings. Moreover, Cluster 10 has a lower standard deviation of the building area, showing more even pattern than Cluster 3. We also include the detailed patterns of each identified cluster in Figure~\ref{composition_figure} (b).

\begin{figure}[h]
\centering
    \includegraphics[trim={0 0 0 0},clip,width=0.5\textwidth]{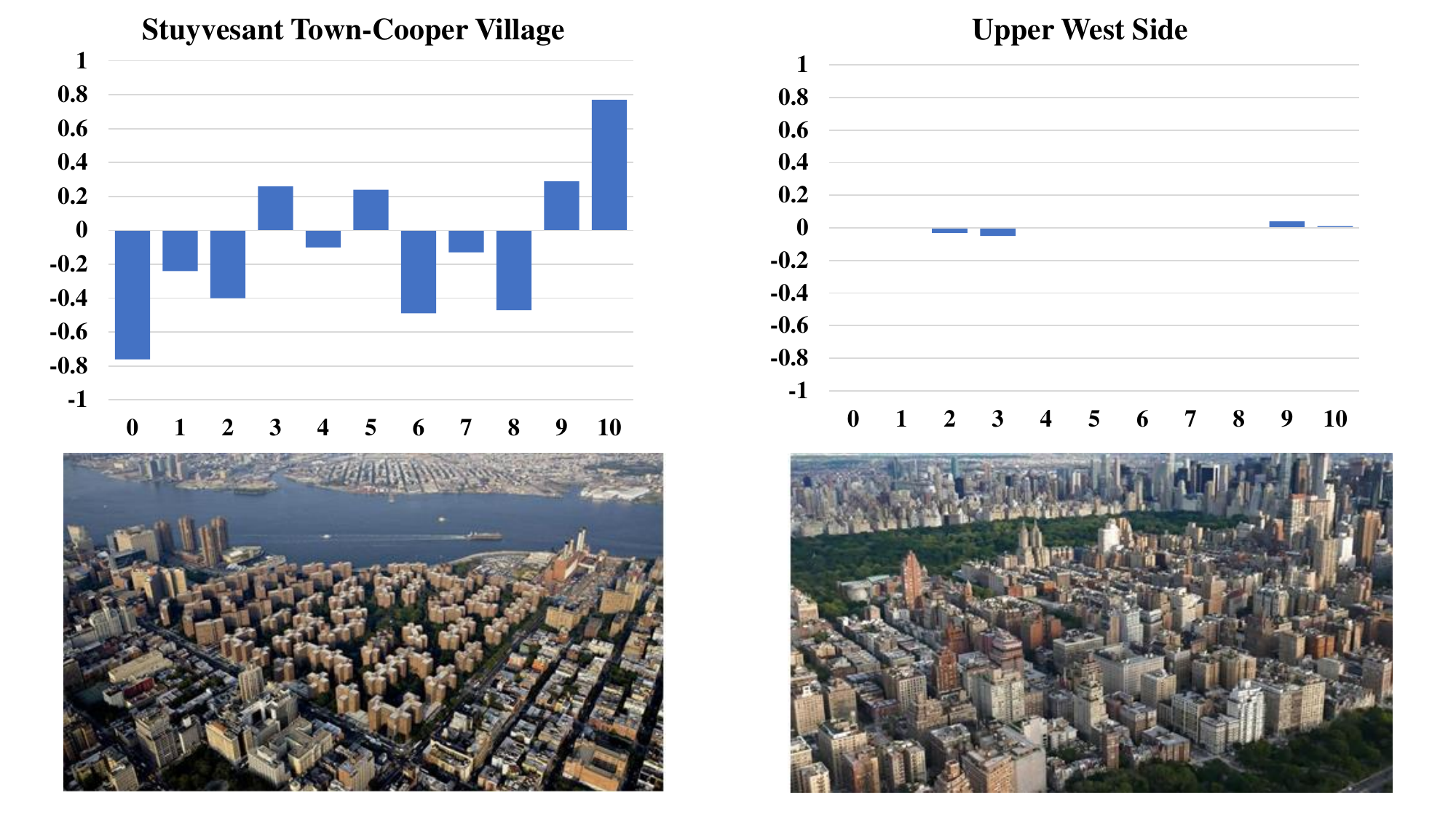}
  \caption{Top row shows deviation of urban layout distributions from the Manhattan baseline and bottom row shows the aerial view of Stuyvesant Town-Peter Cooper Village (left) and the aerial view of Upper West Side (right)
.\label{neiborhood_composition}}
\end{figure}

\textbf{Measurement of neighborhoods' urban layout composition.}
Having identified unique urban layout typologies of Manhattan, our approach allows urban planners to characterize a neighborhood’s urban layout composition. Specifically, we determine the Manhattan’s urban layout composition of each cluster as shown in Figure~\ref{composition_figure}(a), which indicates that Manhattan has a high percentage of cluster 8 and 10 layouts. From Figure~\ref{composition_figure}(b), we can find these two clusters characterize moderate density and even grain - emblematic of the Manhattan grid.

Based on the percentage composition of Manhattan, we further investigate two characteristic neighborhoods' layout composition: Stuyvesant Town-Peter Cooper Village and Upper West Side by calculating the percentage deviation of each cluster from Manhattan baseline. From top row of Figure~\ref{neiborhood_composition}, we can see that the Upper West Side, with its low deviation among clusters, closely resembles the typical typology of Manhattan. In contrast, the deviation in Stuyvesant Town-Peter Cooper Village is significantly higher, indicating a departure from the general characteristics of Manhattan. Two real-world images at the bottom row of Figure~\ref{neiborhood_composition} validate the distinct patterns of these two neighborhoods. 




\section{Conclusions}



In this work, we propose a tree-based deep auto-encoder network, \ourname, to achieve reconstruction and generation of city areal spatial data. Different from previous works that focus on objects and indoor scenes, we put particular emphasis on a large-scale geometric data from a city view. To address the scarcity of city-level datasets, we collect a large-scale real-world building datasets from eight cities that consist of over 3,000,000 geo-referenced building objects. Next, we propose a novel Spatial-Geometric Distance (SGD) metric to organize the raw geometric data to the binary-tree structured data and present a simple but efficient tree-structure neural network for city generation. We employ a binary tree structure, making our model a continuous L-system implemented as a deep network and enables end-to-end learning from scratch. The addition of LSTM Cell enables our model to build a deeper tree structure. Besides, to better quantitatively measure the quality of building generation, we introduce a novel evaluation metric, Overlapping Area Ratio (OAR). Experiments demonstrate the effectiveness of this hierarchical structure on taking advantage of the spatial hierarchy that can be efficiently discovered by hierarchical clustering via data preprocessing. Meanwhile, two urban planning applications using the latent representations learned by \ourname~also verify our method can serve as an effective way for describing the complex geometry of urban layouts which can benefit modern urban planning.



\section{Acknowledgement}
The research is supported by NSF
FW-HTF program under DUE-2026479. The authors gratefully acknowledge Yuqiong Li and Zhiding Yu for collecting the RealCity3D dataset, and the constructive comments and suggestions from the other collaborators.



\bibliography{bibliography}

\begin{thebibliography}{xx}

\bibitem[Abualdenien and Borrmann, 2020]{abualdenien2020}
Abualdenien, J. and Borrmann, A., 2020.
\newblock Formal analysis and validation of levels of geometry (log) in building information models.
\newblock In: 27th International Workshop on Intelligent Computing in Engineering.

\bibitem[Achlioptas et al., 2018]{achlioptas2018learning}
Achlioptas, P., Diamanti, O., Mitliagkas, I. and Guibas, L., 2018.
\newblock Learning representations and generative models for 3d point clouds.
\newblock In: International conference on machine learning, PMLR, pp.~40--49.

\bibitem[Bauchet and Lafarge, 2019]{bauchet2019city}
Bauchet, J.-P. and Lafarge, F., 2019.
\newblock City reconstruction from airborne lidar: A computational geometry approach.
\newblock ISPRS Annals of the Photogrammetry, Remote Sensing and Spatial Information Sciences 4, pp.~19--26.

\bibitem[Chang et al., 2015]{chang2015shapenet}
Chang, A.~X., Funkhouser, T., Guibas, L., Hanrahan, P., Huang, Q., Li, Z., Savarese, S., Savva, M., Song, S. and Su, H. e.~a., 2015.
\newblock Shapenet: An information-rich 3d model repository.

\bibitem[Chang et al., 2021]{chang2021building}
Chang, K.-H., Cheng, C.-Y., Luo, J., Murata, S., Nourbakhsh, M. and Tsuji, Y., 2021.
\newblock Building-gan: Graph-conditioned architectural volumetric design generation.
\newblock arXiv preprint arXiv:2104.13316.

\bibitem[Chen et al., 2018]{chen2018tree}
Chen, X., Liu, C. and Song, D., 2018.
\newblock Tree-to-tree neural networks for program translation.
\newblock In: Advances in neural information processing systems, pp.~2547--2557.

\bibitem[Chu et al., 2019]{chu2019neural}
Chu, H., Li, D., Acuna, D., Kar, A., Shugrina, M., Wei, X., Liu, M.-Y., Torralba, A. and Fidler, S., 2019.
\newblock Neural turtle graphics for modeling city road layouts.
\newblock In: Proceedings of the IEEE International Conference on Computer Vision, pp.~4522--4530.

\bibitem[Demir et al., 2014]{demir2014proceduralization}
Demir, I., Aliaga, D.~G. and Benes, B., 2014.
\newblock Proceduralization of buildings at city scale.
\newblock In: 2014 2nd International Conference on 3D Vision, Vol.~1, IEEE, pp.~456--463.

\bibitem[Etten et al., 2019]{vanetten2019spacenet}
Etten, A.~V., Lindenbaum, D. and Bacastow, T.~M., 2019.
\newblock Spacenet: A remote sensing dataset and challenge series.

\bibitem[Gadelha et al., 2018]{gadelha2018multiresolution}
Gadelha, M., Wang, R. and Maji, S., 2018.
\newblock Multiresolution tree networks for 3d point cloud processing.
\newblock In: Proceedings of the European Conference on Computer Vision (ECCV), pp.~103--118.

\bibitem[Gao et al., 2021]{Gao_2021}
Gao, W., Nan, L., Boom, B. and Ledoux, H., 2021.
\newblock {SUM}: A benchmark dataset of semantic urban meshes.
\newblock {ISPRS} Journal of Photogrammetry and Remote Sensing 179, pp.~108--120.

\bibitem[Goller and Kuchler, 1996]{goller1996learning}
Goller, C. and Kuchler, A., 1996.
\newblock Learning task-dependent distributed representations by backpropagation through structure.
\newblock In: Proceedings of International Conference on Neural Networks (ICNN'96), Vol.~1, IEEE, pp.~347--352.

\bibitem[Groueix et al., 2018]{groueix2018papier}
Groueix, T., Fisher, M., Kim, V.~G., Russell, B.~C. and Aubry, M., 2018.
\newblock A papier-m{\^a}ch{\'e} approach to learning 3d surface generation.
\newblock In: Proceedings of the IEEE conference on computer vision and pattern recognition, pp.~216--224.

\bibitem[Gui and Qin, 2021]{gui2021automated}
Gui, S. and Qin, R., 2021.
\newblock Automated lod-2 model reconstruction from very-high-resolution satellite-derived digital surface model and orthophoto.
\newblock ISPRS Journal of Photogrammetry and Remote Sensing 181, pp.~1--19.

\bibitem[Guo et al., 2020]{guo2020inverse}
Guo, J., Jiang, H., Benes, B., Deussen, O., Zhang, X., Lischinski, D. and Huang, H., 2020.
\newblock Inverse procedural modeling of branching structures by inferring l-systems.
\newblock ACM Transactions on Graphics (TOG) 39(5), pp.~1--13.

\bibitem[Gupta et al., 2021]{gupta2021layouttransformer}
Gupta, K., Lazarow, J., Achille, A., Davis, L.~S., Mahadevan, V. and Shrivastava, A., 2021.
\newblock Layouttransformer: Layout generation and completion with self-attention.
\newblock In: Proceedings of the IEEE/CVF International Conference on Computer Vision, pp.~1004--1014.

\bibitem[Ha and Eck, 2017]{ha2017neural}
Ha, D. and Eck, D., 2017.
\newblock A neural representation of sketch drawings.
\newblock arXiv preprint arXiv:1704.03477.

\bibitem[Hu et al., 2020]{hu2020graph2plan}
Hu, R., Huang, Z., Tang, Y., Van~Kaick, O., Zhang, H. and Huang, H., 2020.
\newblock Graph2plan: Learning floorplan generation from layout graphs.
\newblock ACM Transactions on Graphics (TOG) 39(4), pp.~118--1.

\bibitem[Huang et al., 2022]{huang2022city3d}
Huang, J., Stoter, J., Peters, R. and Nan, L., 2022.
\newblock City3d: Large-scale building reconstruction from airborne lidar point clouds.
\newblock Remote Sensing 14(9), pp.~2254.

\bibitem[Johnson, 1967]{johnson1967hierarchical}
Johnson, S.~C., 1967.
\newblock Hierarchical clustering schemes.
\newblock Psychometrika 32(3), pp.~241--254.

\bibitem[Jyothi et al., 2019]{jyothi2019layoutvae}
Jyothi, A.~A., Durand, T., He, J., Sigal, L. and Mori, G., 2019.
\newblock Layoutvae: Stochastic scene layout generation from a label set.
\newblock In: Proceedings of the IEEE/CVF International Conference on Computer Vision, pp.~9895--9904.

\bibitem[Klokov and Lempitsky, 2017]{klokov2017escape}
Klokov, R. and Lempitsky, V., 2017.
\newblock Escape from cells: Deep kd-networks for the recognition of 3d point cloud models.
\newblock In: Proceedings of the IEEE International Conference on Computer Vision, pp.~863--872.

\bibitem[Li et al., 2015]{li2015tree}
Li, J., Luong, M.-T., Jurafsky, D. and Hovy, E., 2015.
\newblock When are tree structures necessary for deep learning of representations?
\newblock arXiv preprint arXiv:1503.00185.

\bibitem[Li et al., 2017]{li2017grass}
Li, J., Xu, K., Chaudhuri, S., Yumer, E., Zhang, H. and Guibas, L., 2017.
\newblock Grass: Generative recursive autoencoders for shape structures.
\newblock ACM Transactions on Graphics (TOG) 36(4), pp.~1--14.

\bibitem[Li et al., 2019]{li2019grains}
Li, M., Patil, A.~G., Xu, K., Chaudhuri, S., Khan, O., Shamir, A., Tu, C., Chen, B., Cohen-Or, D. and Zhang, H., 2019.
\newblock Grains: Generative recursive autoencoders for indoor scenes.
\newblock ACM Transactions on Graphics (TOG) 38(2), pp.~1--16.

\bibitem[LIFULL~Co., 2015]{LIFULL}
LIFULL~Co., L., 2015.
\newblock Lifull home’s dataset.
\newblock \url{https://www.nii.ac.jp/dsc/idr/lifull/}.

\bibitem[Lin et al., 2022]{lin2022capturing}
Lin, L., Liu, Y., Hu, Y., Yan, X., Xie, K. and Huang, H., 2022.
\newblock Capturing, reconstructing, and simulating: the urbanscene3d dataset.
\newblock In: European Conference on Computer Vision, Springer, pp.~93--109.

\bibitem[Liu et al., 2018]{liu2018floornet}
Liu, C., Wu, J. and Furukawa, Y., 2018.
\newblock Floornet: A unified framework for floorplan reconstruction from 3d scans.
\newblock In: Proceedings of the European conference on computer vision (ECCV), pp.~201--217.

\bibitem[Merrell et al., 2010]{merrell2010computer}
Merrell, P., Schkufza, E. and Koltun, V., 2010.
\newblock Computer-generated residential building layouts.
\newblock In: ACM SIGGRAPH Asia 2010 papers, pp.~1--12.

\bibitem[Mo et al., 2019]{mo2019structurenet}
Mo, K., Guerrero, P., Yi, L., Su, H., Wonka, P., Mitra, N. and Guibas, L.~J., 2019.
\newblock Structurenet: Hierarchical graph networks for 3d shape generation.
\newblock arXiv preprint arXiv:1908.00575.

\bibitem[Nash et al., 2020]{nash2020polygen}
Nash, C., Ganin, Y., Eslami, S. and Battaglia, P.~W., 2020.
\newblock Polygen: An autoregressive generative model of 3d meshes.
\newblock arXiv preprint arXiv:2002.10880.

\bibitem[Nauata et al., 2020a]{nauata2020house}
Nauata, N., Chang, K.-H., Cheng, C.-Y., Mori, G. and Furukawa, Y., 2020a.
\newblock House-gan: Relational generative adversarial networks for graph-constrained house layout generation.
\newblock arXiv preprint arXiv:2003.06988.

\bibitem[Nauata et al., 2020b]{nauata2020housegan}
Nauata, N., Chang, K.-H., Cheng, C.-Y., Mori, G. and Furukawa, Y., 2020b.
\newblock House-gan: Relational generative adversarial networks for graph-constrained house layout generation.

\bibitem[Nishida et al., 2016]{nishida2016interactive}
Nishida, G., Garcia-Dorado, I., Aliaga, D.~G., Benes, B. and Bousseau, A., 2016.
\newblock Interactive sketching of urban procedural models.
\newblock ACM Transactions on Graphics (TOG) 35(4), pp.~1--11.

\bibitem[Nooruddin and Turk, 2003]{fakir2003ray}
Nooruddin, F. and Turk, G., 2003.
\newblock Simplification and repair of polygonal models using volumetric techniques.
\newblock IEEE Transactions on Visualization and Computer Graphics 9(2), pp.~191--205.

\bibitem[{NYC DCP}, 2014]{NYCdata_web}
{NYC DCP}, 2014.
\newblock Nycdata.

\bibitem[{NYC DoITT}, 2019]{nycwebsite}
{NYC DoITT}, 2019.
\newblock {NYC} 3-d building model.
\newblock \url{https://www1.nyc.gov/site/doitt/initiatives/3d-building.page}.

\bibitem[Qi et al., 2017a]{qi2017pointnet}
Qi, C.~R., Su, H., Mo, K. and Guibas, L.~J., 2017a.
\newblock Pointnet: Deep learning on point sets for 3d classification and segmentation.
\newblock In: Proceedings of the IEEE conference on computer vision and pattern recognition, pp.~652--660.

\bibitem[Qi et al., 2017b]{qi2017pointnet++}
Qi, C.~R., Yi, L., Su, H. and Guibas, L.~J., 2017b.
\newblock Pointnet++: Deep hierarchical feature learning on point sets in a metric space.
\newblock arXiv preprint arXiv:1706.02413.

\bibitem[Ritchie et al., 2016]{ritchie2016neurally}
Ritchie, D., Thomas, A., Hanrahan, P. and Goodman, N., 2016.
\newblock Neurally-guided procedural models: Amortized inference for procedural graphics programs using neural networks.
\newblock In: Advances in neural information processing systems, pp.~622--630.

\bibitem[Roy et al., 2020]{roy2020tree}
Roy, D., Panda, P. and Roy, K., 2020.
\newblock Tree-cnn: a hierarchical deep convolutional neural network for incremental learning.
\newblock Neural Networks 121, pp.~148--160.

\bibitem[Sharma et al., 2018]{sharma2018csgnet}
Sharma, G., Goyal, R., Liu, D., Kalogerakis, E. and Maji, S., 2018.
\newblock Csgnet: Neural shape parser for constructive solid geometry.
\newblock In: Proceedings of the IEEE Conference on Computer Vision and Pattern Recognition, pp.~5515--5523.

\bibitem[Socher et al., 2011]{socher2011parsing}
Socher, R., Lin, C.~C., Manning, C. and Ng, A.~Y., 2011.
\newblock Parsing natural scenes and natural language with recursive neural networks.
\newblock In: Proceedings of the 28th international conference on machine learning (ICML-11), pp.~129--136.

\bibitem[Spacenet dataset, n.d.]{spacenetdataset}
Spacenet dataset, n.d.
\newblock \url{https://spacenetchallenge.github.io/}.

\bibitem[{Stadt Zurich}, 2018]{zurichwebsite}
{Stadt Zurich}, 2018.
\newblock {Zurich} 3-d building model.
\newblock \url{https://www.stadt-zuerich.ch/ted/de/index/geoz/geodaten_u_plaene/3d_stadtmodell.html}.

\bibitem[Tai et al., 2015]{tai2015improved}
Tai, K.~S., Socher, R. and Manning, C.~D., 2015.
\newblock Improved semantic representations from tree-structured long short-term memory networks.
\newblock arXiv preprint arXiv:1503.00075.

\bibitem[Vanegas et al., 2012a]{vanegas2012inverse}
Vanegas, C.~A., Garcia-Dorado, I., Aliaga, D.~G., Benes, B. and Waddell, P., 2012a.
\newblock Inverse design of urban procedural models.
\newblock ACM Transactions on Graphics (TOG) 31(6), pp.~1--11.

\bibitem[Vanegas et al., 2012b]{vanegas2012procedural}
Vanegas, C.~A., Kelly, T., Weber, B., Halatsch, J., Aliaga, D.~G. and M{\"u}ller, P., 2012b.
\newblock Procedural generation of parcels in urban modeling.
\newblock In: Computer graphics forum, Vol.~31, Wiley Online Library, pp.~681--690.

\bibitem[Williams and Zipser, 1989]{williams1989learning}
Williams, R.~J. and Zipser, D., 1989.
\newblock A learning algorithm for continually running fully recurrent neural networks.
\newblock Neural computation 1(2), pp.~270--280.

\bibitem[Wu et al., 2019]{wu2019data}
Wu, W., Fu, X.-M., Tang, R., Wang, Y., Qi, Y.-H. and Liu, L., 2019.
\newblock Data-driven interior plan generation for residential buildings.
\newblock ACM Transactions on Graphics (TOG) 38(6), pp.~1--12.

\bibitem[Xu et al., 2021a]{xu2021blockplanner}
Xu, L., Xiangli, Y., Rao, A., Zhao, N., Dai, B., Liu, Z. and Lin, D., 2021a.
\newblock Blockplanner: city block generation with vectorized graph representation.
\newblock In: Proceedings of the IEEE/CVF International Conference on Computer Vision, pp.~5077--5086.

\bibitem[Xu et al., 2021b]{xu2021voxel}
Xu, Y., Tong, X. and Stilla, U., 2021b.
\newblock Voxel-based representation of 3d point clouds: Methods, applications, and its potential use in the construction industry.
\newblock Automation in Construction.

\bibitem[Yanai et al., 2017]{yanai2017poisson}
Yanai, S., Umegaki, R., Hasegawa, K., Li, L., Yamgushi, H. and Satoshi, T., 2017.
\newblock Improving transparent visualization of large-scale laser-scanned point clouds using poisson disk sampling.
\newblock In: 2017 International Conference on Culture and Computing (Culture and Computing), pp.~13--19.

\bibitem[Yang et al., 2013]{yang2013urban}
Yang, Y.-L., Wang, J., Vouga, E. and Wonka, P., 2013.
\newblock Urban pattern: Layout design by hierarchical domain splitting.
\newblock ACM Transactions on Graphics (TOG) 32(6), pp.~1--12.

\bibitem[Zhang et al., 2020]{zhang2020conv}
Zhang, F., Nauata, N. and Furukawa, Y., 2020.
\newblock Conv-mpn: Convolutional message passing neural network for structured outdoor architecture reconstruction.
\newblock In: Proceedings of the IEEE/CVF Conference on Computer Vision and Pattern Recognition, pp.~2798--2807.

\bibitem[Zhao et al., 2022]{zhao2022extracting}
Zhao, W., Persello, C. and Stein, A., 2022.
\newblock Extracting planar roof structures from very high resolution images using graph neural networks.
\newblock ISPRS Journal of Photogrammetry and Remote Sensing 187, pp.~34--45.

\bibitem[Zhou et al., 2020]{zhou2020holicity}
Zhou, Y., Huang, J., Dai, X., Luo, L., Chen, Z. and Ma, Y., 2020.
\newblock Holicity: A city-scale data platform for learning holistic 3d structures.
\newblock arXiv preprint arXiv:2008.03286.

\end{thebibliography}



\end{document}